%% file: main.tex
\pdfoutput=1
\documentclass[a4paper,twocolumn]{article}

\usepackage[T1]{fontenc}
\usepackage{hyperref}
\usepackage{xcolor}

\usepackage{subfig}
\usepackage[pdftex]{graphicx}
\graphicspath{{./figs/}}
\DeclareGraphicsExtensions{.pdf}

\usepackage{amsmath}
\usepackage{amssymb}

\usepackage{algorithm}
\usepackage{algpseudocode}
\usepackage{forest}
\usepackage{multirow}

\usepackage{siunitx}

\usepackage{natbib}

\usepackage{pdfcomment}

\newcommand\cinit[0]{C_{\text{I}}}
\newcommand\cinside[0]{C_{\text{G}}}
\newcommand\Pprev[0]{\mathcal{P}_{\text{prev}}}
\newcommand\cg[0]{\cinside}

\newcommand\bnear[0]{\text{NN}}
\newcommand\bnext[0]{C_{\text{N}}}

\newcommand\crand[0]{\text{RC}}
\newcommand\cfree[0]{\text{CF}}

\newcommand\steered[0]{\mathcal{P}_{\bnear,\crand}}
\newcommand\steeredcf[0]{\mathcal{P}_{\bnear,\cfree}}
\newcommand\steereds[0]{\mathcal{P}_{C_k,\ce}}
\newcommand\steeredscf[0]{\mathcal{P}_{C_k,\cfree}}

\newcommand\cn[0]{C_{\text{M}}}
\newcommand\ce[0]{C_{\text{E}}}
\newcommand\cep[0]{C_{\text{E}}^\text{P}}
\newcommand\cecs[0]{\mathbf{\ce}}
\newcommand\ceps[0]{\mathbf{\cep}}

\usepackage{enumitem}

\usepackage[capitalise]{cleveref}
\crefname{algorithm}{Alg.}{Algs.}

\usepackage{iidtp}
\iidtpTitle{Improving Rapidly-exploring Random Trees algorithm for~Automated Parking in~Real-world Scenarios}
\iidtpAuthors{
\iidtpAuthor{Jiri Vlasak}{https://orcid.org/0000-0002-6618-8152}
\iidtpAuthor{Michal Sojka}{https://orcid.org/0000-0002-8738-075X}
\iidtpAuthor{Zden\v{e}k Hanz\'{a}lek}{https://orcid.org/0000-0002-8135-1296}
}
\iidtpInfo{
\iidtpInfoURL{Source code}{https://rtime.ciirc.cvut.cz/gitweb/hubacji1/iamcar2.git}
\iidtpInfoURL{Video}{https://www.youtube.com/watch?v=u_Vqfd5Cn8Q}
}
\iidtpBibKey{Vlasak26082025}
\iidtpPublisher{ccby}
\iidtpJournal{Journal of Intelligent Transportation Systems}
\iidtpYear{2025}

\begin{document}
\makeiidtp

\title{Improving Rapidly-exploring Random Trees algorithm for~Automated Parking in~Real-world Scenarios}

\author{
	Jiri Vlasak$^1$\footnote{https://orcid.org/0000-0002-6618-8152}
	, Michal Sojka$^2$\footnote{https://orcid.org/0000-0002-8738-075X}
	, and Zden\v{e}k Hanz\'{a}lek$^2$\footnote{https://orcid.org/0000-0002-8135-1296}}

\date{{\small
$^1$Faculty of Electrical Engineering, Czech Technical University in Prague
\\
$^2$Czech Institute of Informatics, Robotics and Cybernetics, Czech Technical University in Prague
\\
\{jiri.vlasak.2, michal.sojka, zdenek.hanzalek\}@cvut.cz
}}

\maketitle

\abstract{
Automated parking is a self-driving feature that has been available in cars for several years now. However, parking assistants in currently produced cars fail to park in more complex scenarios and require the driver to move the car to an expected starting position before the assistant is activated. We overcome these limitations by proposing a planning algorithm consisting of two stages: (1)~a~geometric planner for maneuvering inside the parking slot and (2)~a~Rapidly-exploring Random Trees (RRT)-based planner that finds a collision-free path from the initial position to the slot entry. Evaluation of computational experiments demonstrates that improvements over commonly used RRT extensions reduce the parking path cost by 21\% and reduce the computation time by 79.5\%. The suitability of the algorithm for real-world parking scenarios was verified in physical experiments with Porsche Cayenne.
}

\input{intro}
\input{problem}
\input{psp}
\input{epp}
\input{eval}

\section{Conclusion}

\label{s:concl}
In this paper, we proposed a planning algorithm for automated vehicle parking.  The algorithm can deal with detected obstacles and its output is a path that the vehicle control system, such as the one described in this paper, could follow. The algorithm includes two different planners for different phases of the parking process. The in-slot planner provides the candidate entry position and maneuver inside the parking slot. Then, the RRT-based out-of-slot planner computes a drivable, collision-free, near-optimal path from the initial position to a candidate entry position.

Compared to other RRT-based planners, our algorithm provides the following improvements: (1) the parking position is given by the coordinates of the whole parking slot  instead of just a single point,
(2) the use of different heuristics in the search and build phases of the RRT algorithm,
(3) the reuse of past paths in the Anytime RRT,
(4) the concept of goal zone to check whether the RRT algorithm has reached the goal, and
(5) the path optimization based on Dijkstra's shortest path algorithm.

By using the cost heuristic, the goal zone, and the path optimization, the number of iterations required to achieve a zero error rate is reduced by \SI{74}{\%}. Similarly, the final path cost is improved by \SI{21}{\%} on average and the computation time is improved by \SI{79.5}{\%}. Our algorithm results in paths similar to those human drivers would drive themselves, which has been confirmed in physical experiments with the Porsche Cayenne.

The limitations of our algorithm arise from the fact that the path is composed of Reeds-Shepp curves. That is: the path curvature is discontinuous and uses only the maximum steering angles.
In our future work, we intend to address these limitations by using a different interpolation algorithm for the final path optimization. The RRT algorithm, where each iteration must be fast, will continue to use simple Reeds-Shepp curves. We would also like to improve the performance of our algorithm by replacing uniform sampling with more targeted sampling.

\section*{Acknowledgement}
\label{s:ack}

This work was co-funded by the Grant Agency of the Czech Technical University in Prague under the grant No. SGS22/167/OHK3/3T/13 and by the European Union under the project ROBOPROX (reg. no. CZ.02.01.01/00/22\_008/0004590).
This work contributes to the sustainability of project CZ.02.1.01/0.0/0.0/16\_026/ 0008432 Cluster 4.0 -- Methodology of System Integration, financed by European Structural and Investment Funds and Operational Programme Research, Development and Education via Ministry of Education, Youth and Sports of the Czech Republic.

\section*{Disclosure}
The authors report there are no competing interests to declare.

%\section{References}

\bibliographystyle{plainnat}
\bibliography{main}

% \appendix
% \section{...}

\end{document}

%% file: intro.tex
\section{Introduction}
\label{s:intro}

The automotive industry is currently experiencing the boom of self-driving cars. We still have to wait for fully automated cars, but the level of automation is increasing every year and advanced features such as adaptive cruise control, lane departure warning or parking assistance are available in more and more car models.

The parking assistants currently available in production vehicles suffer from several limitations. They force the driver to stop the car in a certain position with respect to a parking slot, e.g., in case of parallel parking, behind the parking slot with the heading parallel to the curb. They calculate the path of the parking maneuver based on geometric equations and do not take into account the obstacles in more complex environments.

In this paper, we overcome the above limitations by proposing a planning algorithm consisting of two phases that are executed in reverse order to the parking process itself.

\begin{figure}[H]
    \centering
    \includegraphics[width=\linewidth]{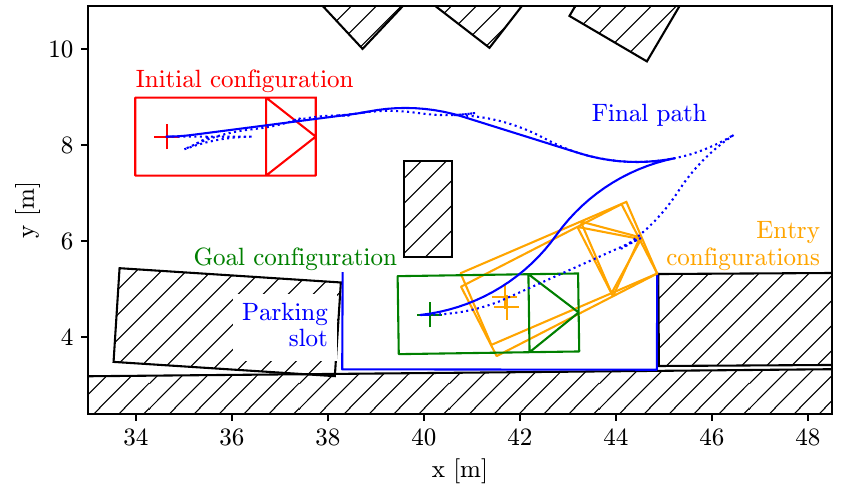}
    \caption{A real-world parking scenario. The black shaded objects are obstacles, the blue frame is the parking slot, the initial position is red, the entry positions are orange, and the goal position is green. The path before optimization is dotted in blue and the optimized final path is solid blue. The cross represents the center of the rear axle.}
    \label{f:intro}
\end{figure}

The phases are:

\begin{enumerate}
    \item \emph{In-slot planning} to find the path inside the slot and a so-called slot entry position. The path is calculated deterministically by a geometric planner.
    \item \emph{Out-of-slot planning} to calculate a collision-free path from an initial car position to the entry position found in the in-slot phase.
\end{enumerate}

% The goal of the in-slot phase is to minimize the number of direction changes. Unlike other planners, our in-slot planner provides not only the path and the final parking position within the parking slot, but also the corresponding entry position, which is then used by the out-of-slot planner. This approach gives us more flexibility to enter the slot.

% Our out-of-slot planner is based on the Rapidly-exploring Random Trees (RRT) algorithm. We have extended it by several enhancements: the use of the nearest neighbor search heuristic, the concept of a goal zone, path optimization based on Dijkstra's shortest path algorithm, and the reset function based on Anytime RRT. Compared to traditional parking assistants, our out-of-slot planner does not require the initial position to be in a specific position relative to the parking slot and can deal with any static obstacles.

The particular contributions of this paper are:

\begin{itemize}
	\item We formulate the parking problem differently from other authors. The goal parking position and the corresponding entry position are not explicitly given as inputs, but are the outputs of our \emph{in-slot planner}. The in-slot planner has the location and size of the entire parking slot as inputs.
	\item We present several improvements to the Rapidly-exploring Random Trees (RRT) algorithm used by our \emph{out-of-slot planner}: i)~use of different metrics in different steps of the algorithm; an inaccurate but fast heuristic to find the nearest nodes, and a slower but accurate computation to find the cost of newly added nodes, ii)~reuse of nodes from the previous iterations of the Anytime RRT algorithm, resulting in finding better paths faster.
	\item We introduce two concepts that are independent of RRT and improve the \emph{out-of-slot planner}: i)~we developed the goal zone concept, which allows us to efficiently evaluate the termination condition of the search; many other authors base termination condition on Euclidean distance from the goal position, which is problematic for non-holonomic vehicles, ii)~we describe path optimization based on Dijkstra's algorithm.
	\item We present the results of extensive computational experiments as well as experiments with a real Porsche Cayenne -- see the video of the physical experiments at \url{https://youtu.be/u_Vqfd5Cn8Q}.
\end{itemize}

This paper is organized as follows. In Section~\ref{s:problem}, we formally define the parking problem. In Section~\ref{s:psp}, we introduce the in-slot planner. We describe enhancements of the RRT algorithm and propose the out-of-slot planner in Section~\ref{s:epp}. In Section~\ref{s:eval}, we evaluate our approach using both computational and physical experiments and discuss the results. The source code of our algorithm is publicly available\footnote{https://rtime.ciirc.cvut.cz/gitweb/hubacji1/iamcar2.git}.

\subsection{Related Work}
\label{s:related}

Automated driving, especially parking, has attracted a lot of attention in the last decade.
Also, \cite{classen_experience_2024} show the increase in the acceptance of automated vehicles.
Despite this, automated parking remains one of the important challenges for Smart Mobility \citep{goumiri_smart_2025}.

The general architecture of automated cars is described by \cite{behere_functional_2015}.
They divide the architecture into three main components: Perception, Decision and Control, and Vehicle Platform.
According to their terminology, the main contribution of our work is in the Decision and Control component.
In addition, we dealt with the Vehicle Platform during the tests with the real vehicle.

Summary of planning algorithms for automated vehicles is presented by \cite{gonzalez_review_2016}. The authors categorize planning algorithms into graph-search planners, sampling-based planners, interpolating curve planners, and planners based on numerical optimization. According to this classification, our out-of-slot planner is a sampling-based planner and our in-slot planner falls into the category of interpolating curve planners (geometric planners).

In terms of automated parking, we divide related work into two groups. First, we compare our in-slot planner with planners used for maneuvering inside the parking slot. The second group deals with planners used outside the parking slot and focuses on sampling-based planners for non-holonomic vehicles.

\subsubsection{In-slot planners}

In-slot planners found in the literature can be divided into two classes: geometric planners \citep{vorobieva_automatic_2015,li_implementation_2016,petrov_geometric_2018} and numerical optimization-based planners \citep{zips_optimisation_2016,li_time-optimal_2016,jing_multi-objective_2018}.

Planners in both classes often do not allow direction changes~\citep{li_implementation_2016,jing_multi-objective_2018} or require the goal position to be specified precisely~\citep{vorobieva_automatic_2015,zips_optimisation_2016,petrov_geometric_2018,jing_multi-objective_2018}, which can lead to unnecessary direction changes.

Optimization-based planners are often slower than geometric planners~\citep{li_time-optimal_2016}. However, with good heuristics, their performance can be sufficient for use in real-time~\citep{zips_optimisation_2016}.

The main difference between our in-slot planner and the above work is that our in-slot planner requires only the vehicle dimensions and the parking slot dimensions as input and not the goal position. The goal position is the output of the algorithm, together with the path inside the slot and the best position for entering the slot. We have described our in-slot planner for parallel parking in more detail in our earlier work~\citep{vlasak_parallel_2022}.

\subsubsection{Out-of-slot planners}

Parking in unpredictable environments requires more sophisticated planning algorithms that can cover a larger area and take obstacles into account. Widely used algorithms for this task include graph-search planners, planners based on numerical optimization, and sampling-based planners. In recent years, planners using neural networks have gained prominence.

Optimization-based planners have the advantage of being able to take into account kinematic, dynamic, or other complex constraints.
However, achieving acceptable performance for real-time applications can present a significant challenge.
Therefore, optimization-based planners use various heuristics and simplifications.
For example, \cite{zips_optimisation_2016} find a tree of discrete landmarks and plan the optimal path only between the landmarks.
Conceptually, their approach is not very different from the RRT algorithm used in this work, since they use a random component in the placement of landmarks.
Their planner has been successfully used by \cite{jang_re-plannable_2020} for fully automated perpendicular parking.
\cite{sun_successive_2022} achieve good performance by using A*-Reshaping algorithm to find an initial approximate path which is then optimized with the help of two simplifications: (i) conversion of non-convex collision avoidance constraints to convex feasible sets around the trajectory and (ii) linearization of non-linear dynamic constrains.
Similarly, \cite{chi_optimization-based_2022} combine hybrid A* graph search algorithm with a numerical optimization approach utilized in a Model Predictive Controller.
In both approaches, the number of forward/backward direction changes is determined statically by the A* component and is not changed later by the numerical optimization.
Our RRT-based approach is more flexible in finding the direction change points.

Combination of optimization-based planners with other planning approaches to find an initial solution is often called ``hierarchical planning''~\citep{yang_sobol_2025, wang_automatic_2023, kuang_hybrid_2023}.
\cite{kuang_hybrid_2023} describe automated parking as a nonlinear optimal control problem where the objective function is the parking maneuver completion time. The parking problem is discretized and transformed intro a nonlinear programming problem. Then, the initial path is found using particle swarm optimization and optimal solution with traditional Lagrange-Newton method.
\cite{wang_automatic_2023} use RRT* with Reeds-Shepp curves for the first stage of planning; they improve the RRT* with innovative collision detection method. The result of the first stage is then used as the initial solution for the nonlinear optimization problem.
\cite{yang_sobol_2025} use Sobol-RRT* -- an RRT* algorithm with random sampling improved by Sobol sequences -- to find an initial solution they later use in a numerical optimization algorithm to plan parking trajectory for a four-wheel steering vehicle.
While the Sobol-RRT* is likely faster than the classical RRT used in our work, the use of non-linear programming to solve the optimal control problem requires significantly more computation than our path optimization based on the Dijkstra's algorithm.

Neural network-based planners suffer from several problems. End-to-end planners, such as that of \cite{sousa_parallel_2022}, are capable of reliably parking the vehicle only within a finite distance (1.5 -- 3\,m) from the parking slot. The planner proposed by \cite{liu_parking_2017} uses a neural network to find a smooth trajectory between two nearby points, but to find a path for longer distances, they still need a strategy to find so-called ``middle/switching points''. Such a strategy can be provided, for example, by random or semi-random sampling.

As can be seen, sampling-based planners that include a random component still have an advantage in planning a path over a longer distance. Therefore, our out-of-slot planner is based on sampling. Sampling-based planners extend the set of sampled positions by attempting to connect to the positions (samples) they randomly generate. They can be tuned with different approaches to sample generation and can use arbitrary methods to connect samples. Although the planners are not deterministic, they are probabilistically complete and can be asymptotically optimal.

A classic sampling-based planning algorithm is the Rapidly-exploring Random Trees (RRT) algorithm~\citep{lavalle_rapidly-exploring_1998}. One of the first applications of RRT in the context of automated vehicles was developed by \cite{kuwata_motion_2008}. Several improvements to the RRT algorithm have been proposed, such as RRT-Connect by \cite{kuffner_rrt-connect_2000}, Dynamic RRT by \cite{ferguson_replanning_2006}, or Anytime RRT by \cite{ferguson_anytime_2006}. \cite{karaman_sampling-based_2011} have proposed RRT* -- an asymptotically optimal variant of the RRT algorithm. Our out-of-slot planner is not asymptotically optimal, since we can also obtain good results with the plain RRT, as we explain later.

RRT is very popular in parking path planning~\citep{wang_two-stage_2017, feng_model-based_2018, dong_knowledge-biased_2020, jhang_forward_2020, qiu_efficient_2024, wang_path_2024}.  The planners described in the above papers differ mainly in the sampling strategy used and the extensions to the basic RRT algorithm. However, they also have many similarities. Since the paths generated by RRT are often crooked, most authors implement path smoothing as post-processing~\citep{feng_model-based_2018, dong_knowledge-biased_2020, jhang_forward_2020, qiu_efficient_2024, wang_path_2024}. Similarly, many authors use Reeds-Shepp curves~\citep{reeds_optimal_1990} at least in part of their algorithm~\citep{wang_two-stage_2017,dong_knowledge-biased_2020,jhang_forward_2020}. Only \cite{dong_knowledge-biased_2020} validate the algorithm on a real vehicle.

In this paper, we propose several novel extensions to the RRT algorithm and show that the result can be successfully applied to a real vehicle.

% The main difference between our out-of-slot planner and the sampling-based planners presented in above is that we use two cost functions, compare the mutual reachability of positions based on the goal zone concept, and use Dijkstra's algorithm for path optimization between selected points on the path. In addition, we solve the maneuvering within the parking slot using our in-slot planner so that we plan the path from the initial position to the entry position rather than to the goal position within the parking slot.

In our previous work~\citep{vlasak_accelerated_2019}, we extended the RRT*-based algorithm in a similar way; this work builds on the results of~\citep{vlasak_accelerated_2019} but uses the plain RRT algorithm. In addition, we extend the parking algorithm with a multiphase approach and a goal zone concept, use an improved search cost heuristic, and apply Dijkstra's optimization more efficiently. We also add more computational experiments to consider real-world scenarios and perform tests with a real vehicle, which is possible thanks to our previous experience~\citep{panamera}.

%% file: problem.tex
\section{The Parking Problem}
\label{s:problem}

In this section, we introduce the used terminology and formally define the parking problem.

\subsection{Definitions}
\label{s:problem:d}

A car \emph{configuration} is a tuple $C=(x, y, \theta, s, \phi)$, where $x$ and $y$ are Cartesian coordinates of the rear axle center in the global coordinate system and $\theta$ is car heading. To simplify our notation, we include the control input to car configuration; $s$ is the control input of the direction $s\in\{-1, +1\}$, and $\phi$ is the steering angle $\phi\in [-\phi_\text{max}, \phi_\text{max}]$.
Car configurations are subject to a discrete kinematic model $C_{k+1} = f(C_k)$, $k\in \mathbb{N}$ where the function $f$ is given by~\cref{e:1}:

%\noteMS{TODO - vektorová rovnice?}{?}

\begin{equation}\label{e:1}
\begin{split}
       x_{k+1}&=x_k+ s_k\cdot \Delta\cdot cos(\theta_k)\\
       y_{k+1}&=y_k+ s_k\cdot \Delta\cdot sin(\theta_k)\\
       \theta_{k+1}&=\theta_k + \frac{s_k\cdot \Delta}{b}\cdot tan(\phi_k),
\end{split}
\end{equation}
where $\Delta \in \mathbb{R^+}$ is a positive constant called the \emph{step distance}.
\cref{f:intro} shows the \emph{initial}, \emph{entry}, and \emph{goal} configurations, which we denote by $\cinit$, $\ce$, and $\cinside$, respectively.

Car \emph{dimensions} is a tuple $D=(w, d_\text{f}, d_\text{r}, b, \phi_\text{max})$, where $w$ is the width of the car, $d_\text{f}$ and $d_\text{r}$ are the distance from the rear axle center to the front and rear of the car, respectively, $b$ is the wheelbase (distance between the front and rear axles), and $\phi_\text{max}$ is the maximum steering angle.

A car \emph{frame} $\mathcal{F}(C)$ is a rectangle given by the car configuration $C$ and the car dimensions $D$. Figure~\ref{f:intro} shows the car frame of the initial configuration in red. The red cross is centered at $x, y$, and the triangle inside the frame shows the car heading $\theta$.

A \emph{search space} $\Omega$ is a set of possible configurations $\Omega=\{C|x\in [x_\text{min}, x_\text{max}], y\in [y_\text{min}, y_\text{max}], \theta \in (-\pi, \pi], s\in\{-1, +1\}, \phi\in [-\phi_\text{max}, \phi_\text{max}\}$ for some $x_\text{min}$, $x_\text{max}$, $y_\text{min}$, and $y_\text{max}$.

An \emph{obstacle} is a convex polygon in $\mathbb R^2$.

A \emph{parking slot} is a rectangle that does not intersect with any obstacle and whose one side (called \emph{entry side} in the following text) is adjacent to the road. The rectangle is defined as a tuple $P = (p, \delta, W, L)$, where the point $p\in \mathbb{R}^2$ is a corner of the rectangle on the entry side, $\delta$ is the direction of the entry side relative to $p$, see \cref{f:ceps}, $W > w$ is the width of the parking slot, which is larger than the width of the car, and the length of the parking slot is $L\ge d_\text{f} + d_\text{r}$.

A \emph{possible entry configuration} is a car configuration $\cep=(x,y,\theta)$ with the following properties: (i) the right front corner of the car frame $\mathcal{F}(\cep)$ coincides with the corner $p$ of the parking slot, (ii) the car heading is between the angle parallel to the entry side and the angle perpendicular to the entry side, i.e., $\delta + \pi \leq \theta \leq \delta + 3/2\pi$, and (iii) the configuration frame $\mathcal{F}(\cep)$ does not intersect with non-entry sides of the parking slot. We denote a set of possible entry configurations as $\ceps$ and we can see an example subset of $\ceps$ in \cref{f:ceps}.

A \emph{scenario} is a tuple $S=(\Omega, \cinit, P, O)$, where $\Omega$ is a \emph{search space} defined above, $\cinit$ is an initial configuration, $P$ is a goal parking slot, and $O$ is a set of obstacles.

A \emph{path} $\mathcal{P}_{C_1, \cn}$ from configuration $C_1$ to $\cn$ is an ordered sequence of configurations $(C_1, C_2, \dots, \cn)$, where $C_{k+1}=f(C_k)$ for $k=1, 2, \dots, M-1$ for a sequence of pairs $((s_1, \phi_1), (s_2, \phi_2), \dots, (s_{M-1}, \phi_{M-1}))$.

A \emph{cost} $\mathcal{C}(C_1, C_2)$ between two configurations $C_1$ and $C_2$ is the length of the Reeds-Shepp curve between these configurations. A \emph{path cost} $\mathcal{C}(\mathcal{P}_{C_1, \cn}) = \sum_\text{k=1}^\text{M-1} \mathcal{C}(C_k, C_{k+1})$.

We say that the configuration $C$ \emph{collides} when $\mathcal{F}(C)$ intersects with any obstacle of $O$.

A \emph{feasible path} is one whose configurations do not collide.

\subsection{Problem Statement}
\label{s:problem:p}

Our goal is to find a feasible path $\mathcal{P}_{\cinit, \cinside}$ with a minimum number of backward-forward direction changes, where $\cinit$ is an initial configuration and the frame $\mathcal{F}(\cinside)$ lies entirely within the parking slot $P$. Among all paths with minimum backward-forward direction changes, we search for the one with the lowest cost $\mathcal{C}(\mathcal{P}_{\cinit, \cinside})$ that can be found within a bounded computation time.

To find the solution more efficiently, we decompose the parking problem into two subproblems:
\begin{enumerate}
\item Finding the path inside the parking slot using a geometric approach. The geometric approach is preferable because the parking slot contains no obstacles and the constraints represented by the boundaries of the parking slot are simple to deal with. In addition to the path, we want to find a set of \emph{entry configuration candidates} $\cecs\subset\ceps$ from which the car can park into the slot with a minimum number of direction changes.
\item The second subproblem is to find a feasible path $\mathcal{P}_{\cinit, \ce}$ from the initial configuration $\cinit$ to an entry configuration $\ce\in\cecs$ with the lowest path cost.
\end{enumerate}

\begin{figure}[H]
\centering
\includegraphics[width=\linewidth]{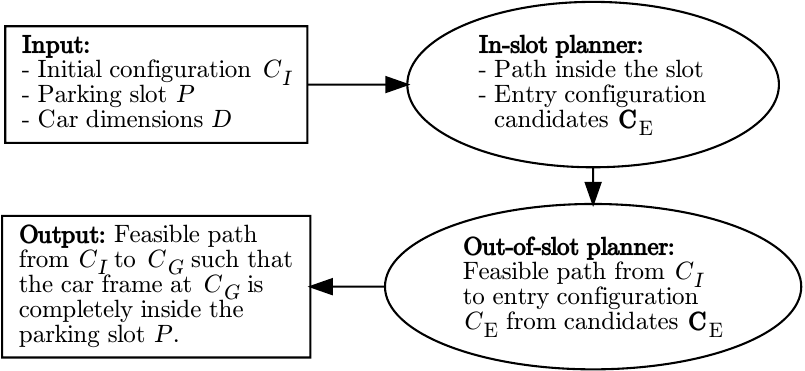}
\caption{Data-flow diagram of our algorithm.}
\label{f:dfd}
\end{figure}

We solve the first subproblem with the so-called in-slot planner, described in \cref{s:psp}, and the second with the out-of-slot planner, discussed in \cref{s:epp}. The data-flow diagram of the overall approach is shown in~\cref{f:dfd}.

%% file: psp.tex
\section{In-slot Planner}
\label{s:psp}

This section describes our planning algorithm for finding a set of entry configuration candidates $\cecs$ to a parking slot $P$ and a corresponding path inside the slot for each $\ce\in\cecs$.

The input to the planner is the car dimensions tuple $D$ and the parking slot $P$% , as defined in Section~\ref{s:problem}
. Compared to related works presented in \cref{s:related}, our in-slot planner is unique in that it provides entry configuration candidates $\cecs$ as output in addition to paths. We use entry configuration candidates as input to our out-of-slot planner described in \cref{s:epp}.

We consider two different cases: Parallel parking and perpendicular parking. Parallel parking requires entering the slot only in backward direction, followed by backward-forward movements. Perpendicular parking, on the other hand, allows entering the slot in both directions and does not require backward-forward movements.

\subsection{Parallel Parking}
\label{s:psp:bfs}

\begin{figure}[H]
\centering
\subfloat[Possible entry configurations. Orange rectangles represent a subset of possible entry configurations.\label{f:ceps}]{
  \includegraphics[width=0.46\linewidth]{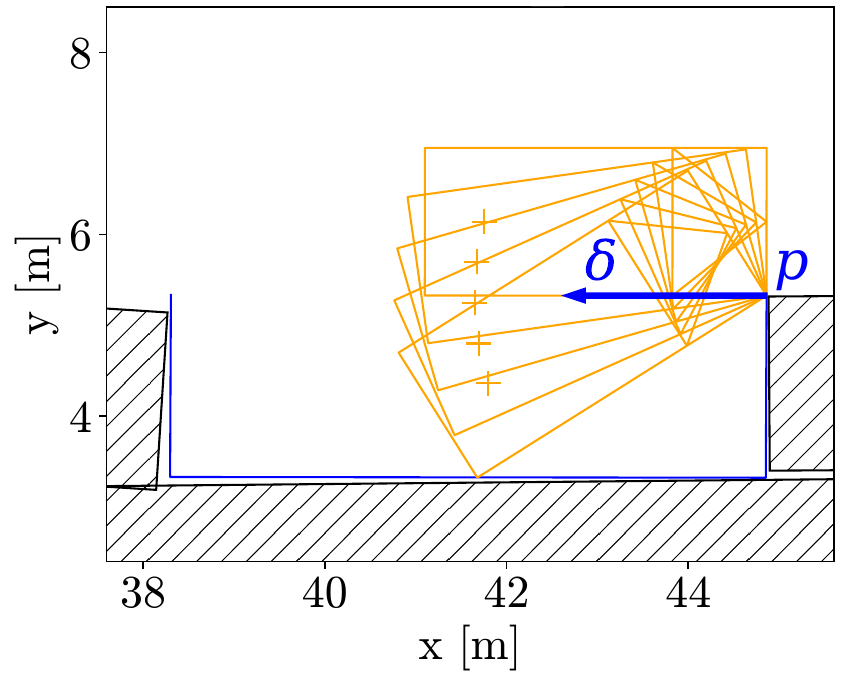}
}
\hfill
\subfloat[Entry and goal configuration candidates. Green rectangles are goal configuration candidates computed by our in-slot planner. Orange rectangles are corresponding entry configuration candidates.\label{f:cecs}]{
  \includegraphics[width=0.46\linewidth]{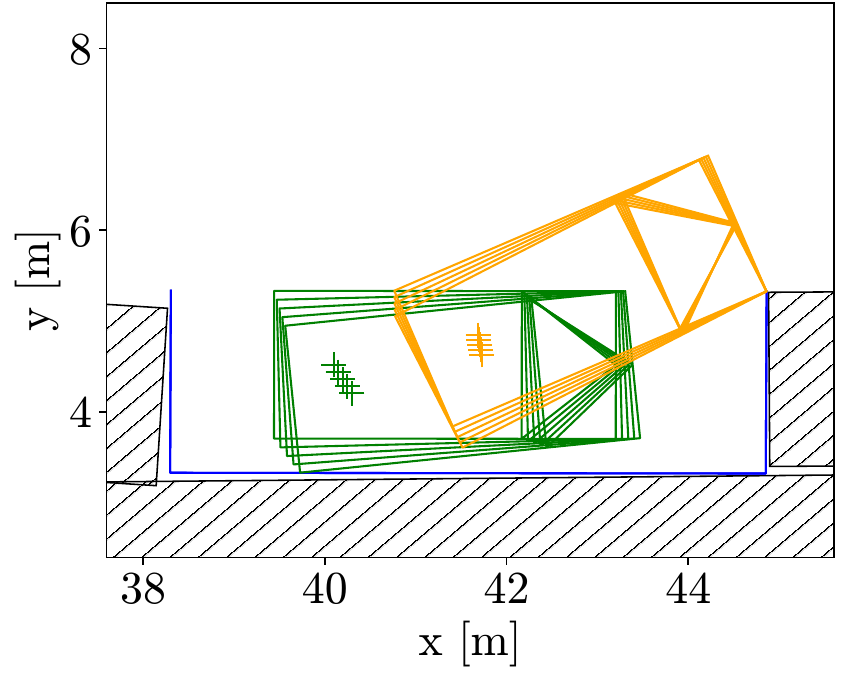}
}
\caption{In-slot planner for parallel parking.}
\end{figure}

Our in-slot planner for parallel parking uses an approach similar to the ``several reversed trials'' planner of \cite{vorobieva_automatic_2015}. Their planner assumes that the goal configuration $\cg$ is known and starts with that configuration. Our in-slot planner starts with a set of possible entry configurations $\ceps$, some of them are shown as orange rectangles in \cref{f:ceps}. For each possible entry configuration $\cep\in\ceps$, the in-slot planner simulates backward-forward movements with the maximum steering angle to find candidates for the goal configuration, shown as green rectangles in \cref{f:cecs}.

The result of the in-slot planner for parallel parking is a series of paths from each entry configuration $\ce\in\cecs$, shown as orange rectangles in \cref{f:cecs}, to the corresponding goal configuration candidate.

\subsection{Perpendicular Parking}
\label{s:psp:perpendicular}

\begin{figure}[H]
\centering
\subfloat[Forward parking.]{
    \begin{tabular}{cc}
    \includegraphics[width=0.46\linewidth]{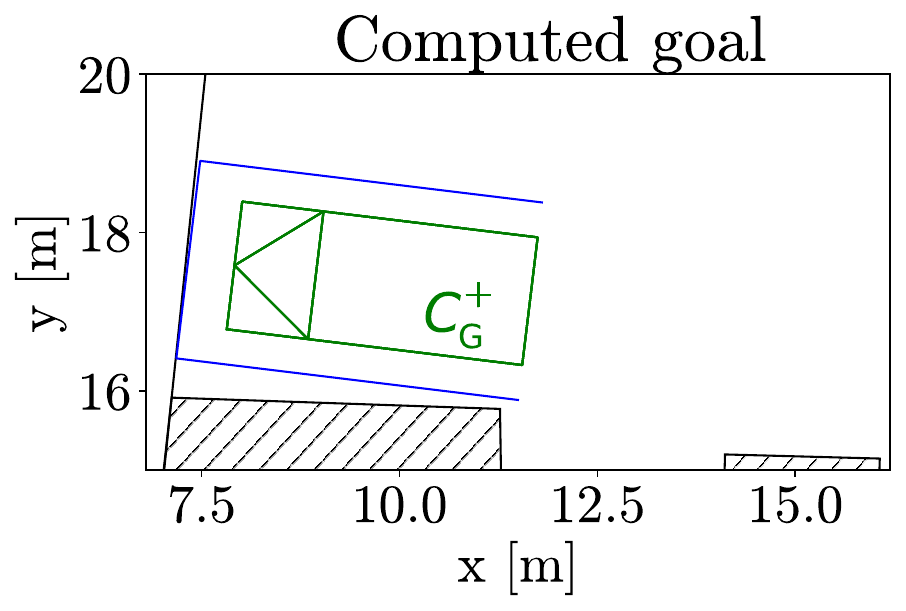} &
    \includegraphics[width=0.46\linewidth]{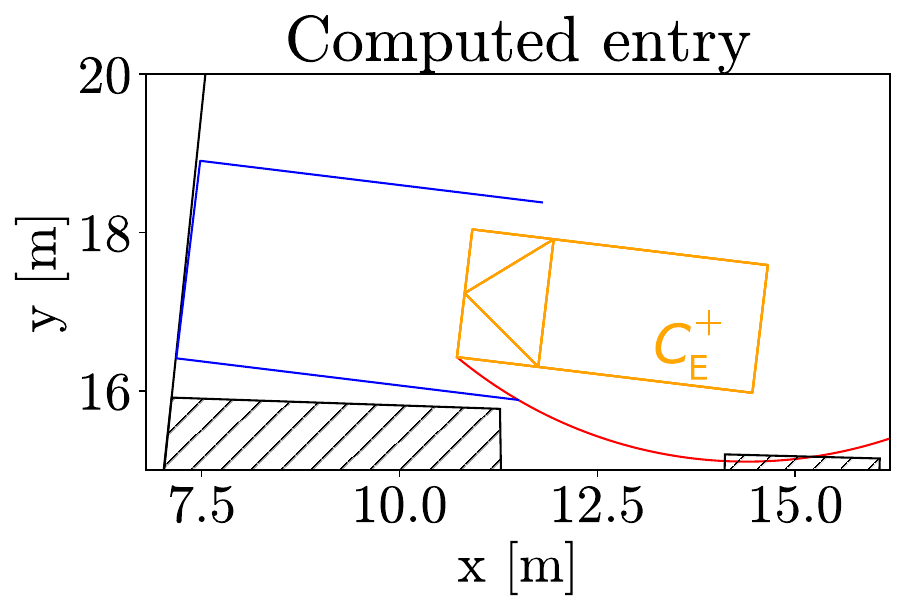}
    \end{tabular}
}
\hfill
\subfloat[Backward parking.]{
    \begin{tabular}{cc}
    \includegraphics[width=0.46\linewidth]{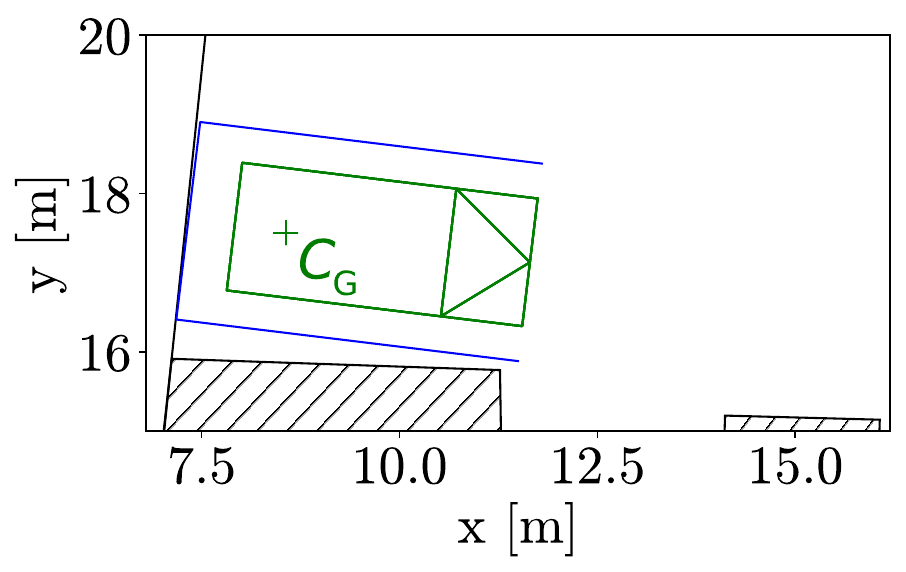} &
    \includegraphics[width=0.46\linewidth]{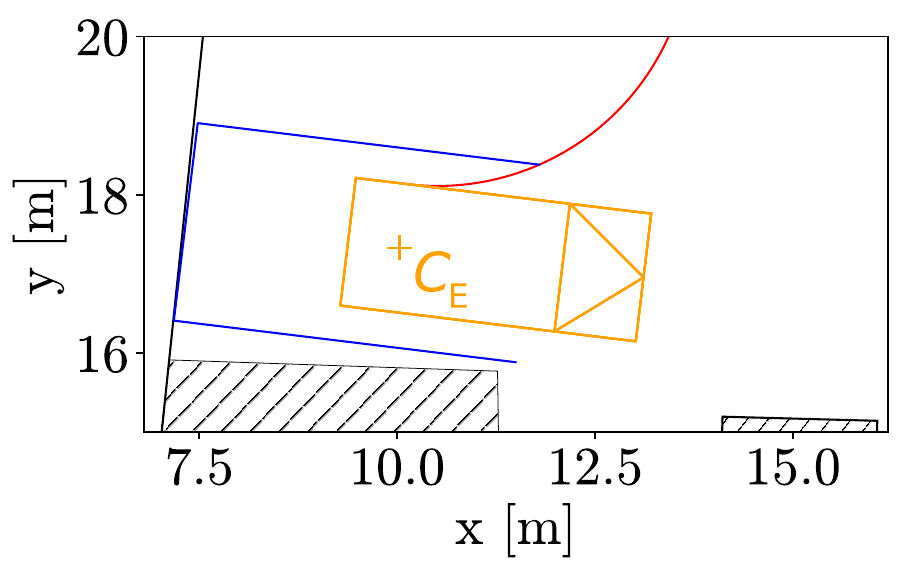}
    \end{tabular}
    \label{f:pe:f}
}
\caption{Forward (backward) perpendicular parking. On the left, there is computed goal configuration for the forward (backward) perpendicular parking. On the right, there is corresponding entry configuration. Red arcs represent path of front (rear) vehicle corners when moving backward (forward) with the maximum steering.}
\label{f:pe}
\end{figure}

In perpendicular parking, the in-slot planner computes one entry configuration for the forward direction and one for the backward direction. Our in-slot planner first computes the goal configuration $\cinside$, shown as a green rectangle in \cref{f:pe}, such that for the forward (backward) direction, the rear (front) side of $\mathcal{F}(\cinside)$ coincides with the entry side of the parking slot $P$. The entry configuration $\ce$, shown as an orange rectangle in \cref{f:pe}, is computed by driving the car with the configuration $\cinside$ out of the parking slot $P$ until the car can leave the slot with maximum steering.

The result of the in-slot planner for the perpendicular parking slot is the path $\mathcal{P}_{\ce, \cg}$ from $\ce$ to $\cg$ for both directions.

%% file: epp.tex
\section{Out-of-slot Planner}
\label{s:epp}

Our out-of-slot planner for finding a feasible path $\mathcal{P}_{\cinit,\ce}$ from the initial to the entry configuration is based on the RRT algorithm extended with Dijkstra-based optimization. The pseudocode of our planner is in \cref{a:osp}.  There, the \textsc{FindPathRRT} function implements the RRT-based algorithm. We recall the original RRT algorithm in~\cref{s:rrt} and describe our improved version (\textsc{FindPathRRT}) in \cref{s:fp}. The call to \textsc{Optimize} improves the path by skipping some configurations, as detailed in \cref{s:opt}. Both steps run several times in a loop, trying to improve the path cost while taking the advantage from the path found in the previous iterations. The algorithm stops when it reaches the maximum number of iterations $I_{max}$ or the path returned by \textsc{FindPathRRT} has not improved in five consecutive iterations. We describe the implementation details in \cref{s:rrt:i}.

\begin{algorithm}[h]
\footnotesize
\caption{Out-of-slot planner}
\label{a:osp}
\begin{itemize}
\item Input: $\cinit$, entry conf. candidates $\cecs$, obstacles $O$.
\item Output: Feasible path $\mathcal{P}_{\cinit,\ce}$ from $\cinit$ to $\ce\in\cecs$.
\end{itemize}
\begin{algorithmic}[1]
\State $\mathcal{P}_{\cinit,\ce}\gets()$\Comment{no path, therefore $\mathcal{C}(\mathcal{P}_{\cinit, \ce})\gets\infty$}
\State $noImprovementCounter\gets 0$
\State $i\gets 0$\Comment{number of iterations}
\While{$i < I_\text{max} \wedge noImprovementCounter < 5$}
  \State $\Pprev \gets \mathcal{P}_{\cinit, \ce}$
  \State $path, i\gets$\Call{FindPathRRT}{$\cinit, \cecs, O, i, \Pprev$}
  \If{$\mathcal{C}(path) < 1.25\cdot\mathcal{C}(\mathcal{P}_{\cinit, \ce})$}
  \State $path \gets$\Call{Optimize}{$path$}\label{a:rrtt:opt}
  \EndIf
  \If{$\mathcal{C}(path) < \mathcal{C}(\mathcal{P}_{\cinit, \ce})$}
  \If{$\mathcal{C}(path) < 0.75\cdot\mathcal{C}(\mathcal{P}_{\cinit, \ce})$}
      \State $noImprovementCounter\gets 0$
    \EndIf
    \State $\mathcal{P}_{\cinit,\ce}\gets path$
  \Else
    \State Increment $noImprovementCounter$
  \EndIf
\EndWhile
\State \textbf{return} $\mathcal{P}_{\cinit,\ce}$
\end{algorithmic}
\end{algorithm}

Compared to the RRT* algorithm~\citep{karaman_sampling-based_2011}, the RRT algorithm is not asymptotically optimal. However, when extending it with the improvements introduced later in this section, the costs of the resulting paths are the same as when using RRT*, but the performance is slightly better due to omitting the rewire procedure of the RRT*.

\subsection{Original RRT}
\label{s:rrt}

Rapidly-Exploring Random Trees (RRT)~\citep{lavalle_rapidly-exploring_1998} is a randomized algorithm for searching a state space. The RRT-based algorithms can be used for solving path planning problems, such as our parking problem defined in \cref{s:problem:p}.

The RRT algorithm uses a tree data structure. The algorithm can handle nonholonomic constraints, arbitrary obstacles, and constraints on configurations. Therefore, the RRT is suitable for solving real-world scenarios.

We denote the tree data structure as $\mathcal{T} = (\cinit, V, E)$, where $\cinit$ is the root of the tree and $V$ is a set of configurations. The set of edges $E$ maintains the parent-child relationship between the configurations such that $(C_i, C_j)\in E \Leftrightarrow C_i, C_j\in\mathcal{P}_{\cinit, \cn}=(\cinit, \dots, C_i, C_j, \dots, \cn)$.
We can see the pseudocode of the original RRT in \cref{a:rrt}.

\begin{algorithm}[h]
\footnotesize
\caption{Original RRT}
\label{a:rrt}
\begin{itemize}
\item Input: $\cinit$, obstacles $O$.
\item Output: $\mathcal{T}$ data structure representing feasible paths between configurations $C\in\Omega\cap\mathcal{T}$
\end{itemize}
\begin{algorithmic}[1]
\State Initialize the tree $\mathcal{T} = (\cinit, \emptyset, \emptyset)$
\While{number of iterations $< I_\text{max}$}\label{a:rrt:bb}
  \State $\crand\gets\Call{RandomConfiguration}{ }$
  \State $\bnear\gets \Call{NearestNeighbor}{\mathcal{T}, \crand}$\label{a:rrt:nn}
  \State $\bnext\gets\Call{Steer}{\bnear,\crand}$
  \If{$\Call{Collides}{\bnext, O}$}
    \State \textbf{continue} the while loop
  \EndIf
  \State \Call{Connect}{$\mathcal{T}, \bnear, \bnext$} \label{a:rrt:ip}
\EndWhile
\end{algorithmic}
\end{algorithm}

% The following basic procedures are used to build the tree $\mathcal{T}$.
% \textsc{RandomConfiguration()} returns random configuration $RC$ from the search space $\Omega$.
% \textsc{NearestNeighbor}$(\mathcal{T}, \crand)$ searches for a configuration $\bnear\in V$ with the lowest cost $\mathcal{C}(\bnear, \crand)$.
% % \textsc{NearNodes}$(\mathcal{T}, \bnear, d)$ returns $\bnear$'s neighboring configurations $\nns$, where $\forall n\in\nns:\mathcal{C}(n,\bnear)<d$.
% \textsc{Steer}$(\bnear,\crand)$ returns the first configuration $\bnext$ of a path $\steered$ from $\bnear$ to $\crand$ that is not in the tree $\mathcal{T}$.
% \textsc{Collides}$(\bnext, O)$ returns whether configuration $\bnext$ collides with an obstacle in $O$ or not.
% \textsc{Connect}$(\mathcal{T}, \bnear, \bnext)$ appends the new configuration $\bnext$ to the tree $\mathcal{T}$, i.e. $V\gets V\cup\bnext$ and $E\gets E\cup(\bnear, \bnext)$.

\subsection{RRT Enhancements}
\label{s:rrtt}
\label{s:fp}

In this section, we describe our enhancements to the original RRT algorithm that allow us to find a path between $\cinit$ and $\cecs$ more efficiently. The pseudocode of the RRT-based \textsc{FindPathRRT} procedure can be found in \cref{a:fp}. Individual enhancements are described in the following subsections.

\begin{algorithm}[h]
\footnotesize
\caption{RRT for finding a path}
\label{a:fp}
\begin{algorithmic}[1]
\Procedure{FindPathRRT}{$\cinit, \cecs, O, i, \Pprev$}
\State Initialize the tree $\mathcal{T} = (\cinit, \emptyset, \emptyset)$
\While{$i < I_\text{max}$}\label{a:rrtt:bb}
  \State $i\gets i + 1$
  \If{$i==1$}
    \State $\crand\gets\ce$
  \Else
    \State $\crand\gets\Call{RandomConfiguration}{\Pprev}$
  \EndIf
  \State $\bnear\gets \Call{NearestNeighbor}{\crand}$\label{a:fp:nn}
  \State $\steered\gets\Call{SteerPath}{\bnear,\crand}$
  \State $\steeredcf\gets\Call{RemoveColliding}{\steered, O}$
  \If{$\steeredcf=\emptyset$}
    \State \textbf{continue} the while loop
  \EndIf
  \State$\Call{Connect}{\mathcal{T},\steeredcf}$\label{a:rrtt:ip}
  \State \Call{SteerTowardEntry}{$\steeredcf, \mathbf\ce$}
  \If{some $\ce\in\mathbf\ce$ added to $\mathcal{T}$}
    \State Find $\mathcal{P}_{\cinit, \ce}$ in $\mathcal{T}$\label{a:rrtt:pf}
    \State \textbf{return} $\mathcal{P}_{\cinit, \ce}, i$
  \EndIf
\EndWhile
\EndProcedure
\end{algorithmic}
\end{algorithm}

\subsubsection{Reuse of Previous Paths with Anytime RRT}
\label{sec:reuse-past-paths}

Our algorithm uses the random sampling procedure shown in \cref{a:rc}, which is inspired by the Anytime RRT~\citep{ferguson_anytime_2006} algorithm. However, instead of repeating random sampling when $\crand_\text{cost}^1 + \crand_\text{cost}^2 >\mathcal{C}(\Pprev)$, we return a randomly chosen configuration from the previously found path $\Pprev$, likely leading to improvement of the current path.

\begin{algorithm}[h]
\footnotesize
\caption{Random sampling}
\label{a:rc}
\begin{algorithmic}[1]
\Procedure{RandomConfiguration}{$\Pprev$}
  \State $\crand\gets$ random sample from $\Omega$
  \State $\crand_\text{cost}^1\gets \Call{SearchCost}{\cinit, \crand}$
  \State $\crand_\text{cost}^2\gets \Call{SearchCost}{\crand, \ce}$
  \If{$\crand_\text{cost}^1 + \crand_\text{cost}^2 >\mathcal{C}(\Pprev)$}
    \State $\crand\gets$ random configuration from  $\Pprev$
  \EndIf
  \State \textbf{return} $\crand$
\EndProcedure
\end{algorithmic}
\end{algorithm}

\subsubsection{Cost Heuristic for Nearest Neighbor Search}
\label{s:fp:sc}

The original RRT algorithm (\cref{a:rrt}) uses the cost function $\mathcal{C}$ in two places: (i) within the \textsc{NearestNeighbor} search (\cref{a:rrt:nn}.) and (ii) when connecting new nodes (\cref{a:rrt:ip}).

From the analysis of the original RRT, we see that it computes the cost $\mathcal{C}$ more often when searching the tree than when connecting.
It appears that the difference in the number of these cases increases with the size of the tree.
If we manage to speed up the cost calculation in \textsc{NearestNeighbor}, we can improve the total computation time significantly.

Therefore, our \textsc{NearestNeighbor} search, uses the K-d tree data structure and computes the cost approximately by applying the heuristic $\mathcal{C}'$ in \cref{e:sc}, which we found to be most suitable for our purposes. $\mathcal{C}'$ estimates the cost of path between configurations $C_\mathcal{T}=(x_\mathcal{T}, y_\mathcal{T}, \theta_\mathcal{T}) \in \mathcal{T}$ and $C_\text{RC}=(x_\text{RC}, y_\text{RC}, \theta_\text{RC})$ as:

\begin{align}
  \mathcal{C}' = \max\left(\Delta_\text{E}, \Delta_\text{A}\cdot \frac{b}{\tan \phi_\text{max}}\right) + 0.1\cdot\Delta_\text{BF},\label{e:sc}
\end{align}
where $b$ and $\phi_{\max}$ are given by car dimensions and the other elements are the following simple functions: $\Delta_\text{E}= \sqrt{(y_\text{RC} - y_\mathcal{T})^2 + (x_\text{RC} - x_\mathcal{T})^2}$ is the Euclidean distance, $\Delta_\text{A}= \min(|\theta_\text{RC} - \theta_\mathcal{T}|, 2\cdot\pi - |\theta_\text{RC} - \theta_\mathcal{T}|)$ is the angular difference, and $\Delta_\text{BF}$ is the cumulative number of backward-forward directions changes between $\cinit$ and $C_\mathcal{T}$.

We use the cost heuristic only when searching the tree for the nearest neighbor in \cref{a:fp:nn} in \cref{a:fp}. We use the length of the Reeds-Shepp curve, which satisfies the kinematic constraints when we connect new nodes in \cref{a:rrt:ip}. Therefore, the paths in the tree also satisfy the kinematic constraints.

\subsubsection{Steering and Collision Checking}
\label{s:fp:c}

The original RRT algorithm extends the tree only by one configuration per iteration and therefore the tree grows slowly. Moreover, in the original RRT algorithm, the tree grows uniformly in all directions, but for our problem, it is advantageous if the growth is directed toward the goal configurations $\mathbf\ce$.

To make the tree grow faster, we extend it by multiple configurations in one iteration by replacing the  \textsc{Steer} procedure with \textsc{SteerPath}, which returns a path between the random configuration and its nearest node in the tree. All the configurations in the path before the first obstacle (if any, see \cref{a:c}) are added to the tree.
% Since the path $\steered$ may collide with obstacles, we replace the \textsc{Collides} function from the original RRT with \textsc{RemoveColliding}, which shortens the path  $\steered$ to the non-colliding subsequence $\steeredcf$, so that $\forall C\in\steeredcf: \neg\textsc{Collides}(C,O)$; it may return the empty path.

\begin{algorithm}[h]
\footnotesize
\caption{Collision check}
\label{a:c}
\begin{algorithmic}[1]
\Procedure{RemoveColliding}{$\steered, O$}
  \State $\cfree\gets\bnear$
  \For{$\bnext\in\steered$}
    \If{\Call{Collides}{$\bnext, O$}}
    \State \textbf{return} $\steeredcf$
    \EndIf
    \State $\cfree\gets\bnext$
  \EndFor
\EndProcedure
\end{algorithmic}
\end{algorithm}

To direct the growth of the tree toward the entry configurations $\cecs$, we try to construct Reeds-Shepp path from each added configuration to an entry configuration and add all their configurations before the potential collision. This is implemented in \textsc{SteerTowardEntry} as detailed in \cref{a:st}. This improvement was originally proposed by~\cite{kuwata_motion_2008}.

\begin{algorithm}[h]
\footnotesize
\caption{Steering toward the entry configuration}
\label{a:st}
\begin{algorithmic}[1]
\Procedure{SteerTowardEntry}{$\steeredcf, \mathbf\ce$}
  %\State$\ce^\text{avg}\gets\Call{AverageConfiguration}{\cecs}$
  \State$\ce^\text{avg}\gets\text{ average of }\cecs$
  \For{$C_k\gets\steeredcf$}\label{a:rrtt:eb}
    \State $\steereds\gets\Call{SteerPath}{C_k,\ce^\text{avg}}$
    \State $\steeredscf\gets\Call{RemoveColliding}{\steereds, O}$
    \If{$\steeredscf=\emptyset$}
      \State \textbf{continue} the for loop
    \EndIf
    \State $\Call{Connect}{\mathcal{T},\steeredscf}$\label{a:rrtt:sc2}    \If{$\cfree\in\Call{$\mathbf{G}$}{\cecs,\theta_{CF}} \wedge |\cfree - \ce| \leq \Delta$}\label{a:rrtt:gf}
      \State Add corresponding $\ce$ to the tree $\mathcal{T}$
    \EndIf
  \EndFor\label{a:rrtt:ee}
\EndProcedure
\end{algorithmic}
\end{algorithm}

\subsubsection{Goal Zone}
\label{s:epp:gz}
\label{s:fp:gz}

To decide whether we have found the complete path, i.e., whether the constructed tree $\mathcal{T}$ contains an entry configuration $\ce\in\mathbf\ce$, we introduce the concept of \emph{goal zone} and use it in \cref{a:rrtt:gf} in \cref{a:st}.

\begin{figure}[h]
    \centering
    \includegraphics[width=\linewidth]{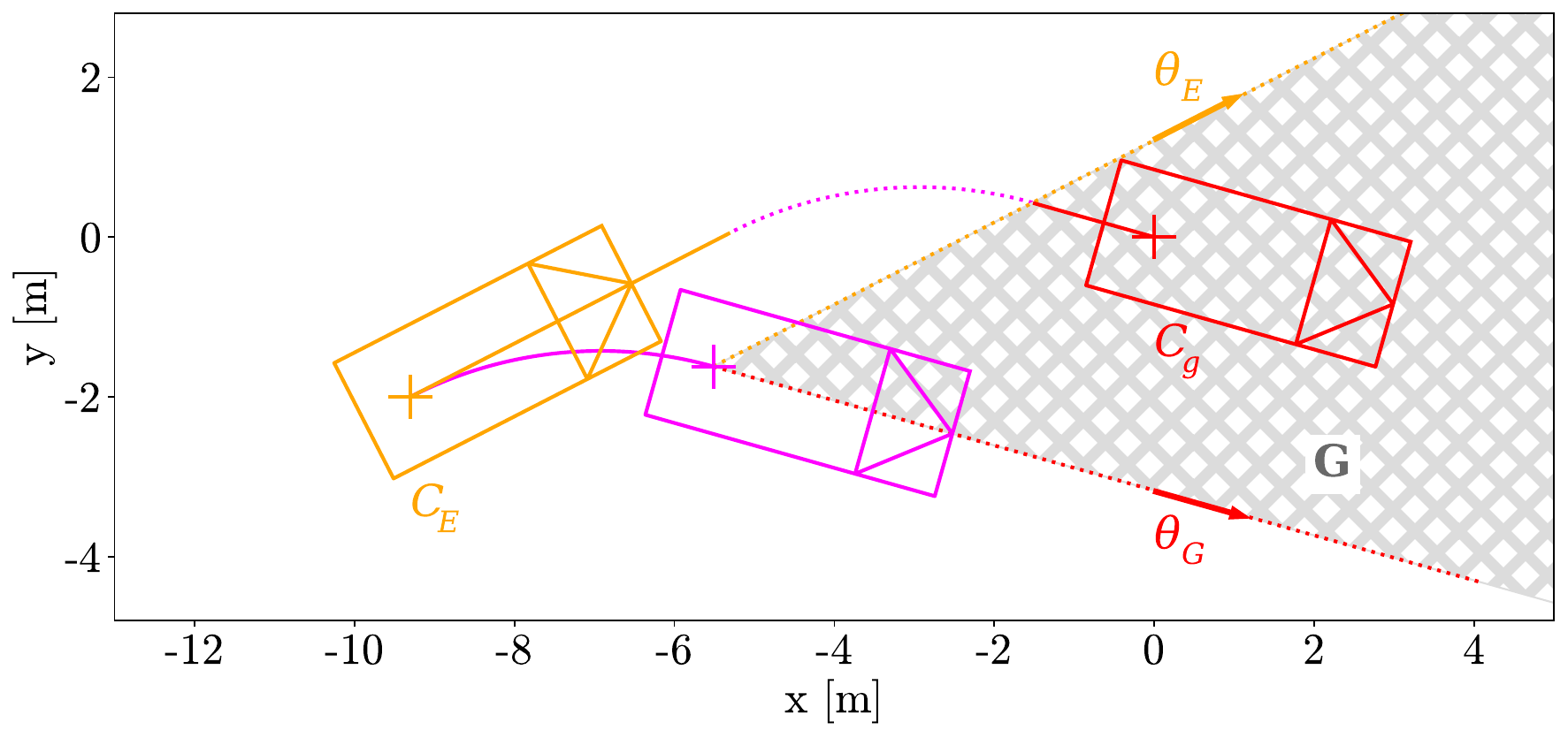}
    \caption{Goal Zone $\mathbf{G}$. $\ce$ is trivially reachable from $C_g$.}
    \label{f:gz}
\end{figure}

The \emph{goal zone} $\mathbf{G}$ of an entry configuration $\ce$ and heading $\theta_G$ is the set of configurations $C_g$, where $\theta_g = \theta_G$ and the configuration $\ce$ is trivially reachable from $C_g$ by a line segment, an arc with maximum steering radius, and a subsequent line segment. Furthemore, we consider the goal zone to be empty if $|\theta_G - \theta_E| \geq \frac{\pi}{2}$.  We see an example of a goal zone $\mathbf{G}$ in \cref{f:gz}.

\subsection{Path Optimization}
\label{s:epp:opt}
\label{s:opt}

It is well known that the RRT algorithm is unlikely to converge to the optimal solution. Although the RRT* is asymptotically optimal, it converges to the optimal solution very slowly. Depending on the scenario, a feasible path returned by the algorithms may be hundreds of percent away from the optimum.

To accelerate convergence to the optimal solution, we introduce path optimization based on Dijkstra's shortest path algorithm. The path optimization algorithm is applicable to both RRT and RRT*. Recall that we use RRT.

The optimization works as follows. Given a path $\mathcal{P}_{\cinit, \ce}$, we find a set of its \emph{interesting configurations}. An \emph{interesting configuration} $C_i$ is either (i) first or last configuration in the path, (ii) a configuration with a change in the direction of motion, or (iii) given an interesting configuration $C_i$, the last configuration $C_j$ in its goal zone is also an interesting configuration, i.e., $C_{j}\in\mathbf{G}(C_i, \theta_j) \wedge C_{j+1}\notin \mathbf{G}(C_i,\theta_{j+1})$ where $i < j$ and $j$ is the smallest value satisfying the condition.

We run Dijkstra's algorithm on the directed graph, where the vertices of the graph represent the interesting configurations. Edges are created between two interesting configurations $C_i, C_j$ if and only if $i<j \wedge C_j\in \mathbf{G}(C_i,\theta_j) \wedge \textsc{SteerPath($C_i$, $C_j$)}\text{ is feasible}$. The cost of the edge is $\mathcal{C}(C_i, C_j)$. Running Dijkstra's algorithm finds a new feasible path from $\cinit$ to $\ce$, possibly skipping some \emph{interesting configurations}.

We perform path optimization in both directions, as we can see in \cref{a:opt}. An example of a path before and after optimization can be found in~\cref{f:intro}.

\begin{algorithm}[h]
\footnotesize
\caption{Path optimization}
\label{a:opt}
\newcommand\interesting[0]{\mathbf I}
\begin{algorithmic}[1]
\Procedure{Optimize}{$\mathcal{P}_{\cinit,\ce}$}
  \While{$\mathcal{C}(\mathcal{P}_{\cinit,\ce})$ is improving}
    \State$\interesting\gets\Call{InterestingConfigurations}{\mathcal{P}_{\cinit,\ce}}$
    \State Run Dijkstra on $\interesting$ starting from $\cinit$
    \State Update $\mathcal{T}$ with the new feasible path $\mathcal{P}_{\cinit,\ce}$
    \State$\mathcal{P}_{\ce,\cinit}\gets\text{ reverse }\mathcal{P}_{\cinit,\ce}$
    \State$\interesting\gets\Call{InterestingConfigurations}{\mathcal{P}_{\ce,\cinit}}$
    \State Run Dijkstra on $\interesting$ starting from $\ce$
    \State Update $\mathcal{T}$ with the new feasible path $\mathcal{P}_{\cinit,\ce}$
  \EndWhile
  \State \textbf{return} $\mathcal{P}_{\cinit,\ce}$
\EndProcedure
\Procedure{InterestingConfigurations}{$\mathcal{P}_{C_1,C_2}$}
  \State $C_i\gets C_1$
  \While{$C_i\neq C_2$}
    \State $\interesting\gets(C_i)$
    \State Find the first $C_j:j>i\wedge C_{j+1}\notin\mathbf{G}(C_i, \theta_{j+1})$
    \State $\interesting\gets\interesting\cup (C_k|i<k<j\wedge s\text{ of }C_k\text{ has changed})$
    \State $\interesting\gets\interesting\cup (C_j)$
    \State $C_i\gets C_j$
  \EndWhile
  \State $\interesting\gets\interesting\cup (C_2)$
  \State \textbf{return} $\interesting$
\EndProcedure
\end{algorithmic}
\end{algorithm}

\subsection{Implementation}
\label{s:rrt:i}

In this section, we provide information about the implementation.

We generate random configurations in \textsc{RandomConfiguration} by selecting it with uniform distribution from the circular subspace $\Omega_\text{C}\subseteq \Omega$. $\Omega_\text{C}=\{C|(x-x_\text{C})^2 + (y-y_\text{C})^2\leq R^2,  \theta\in (-\pi, \pi]\}$, where $x_\text{C}$ and $y_\text{C}$ are Cartesian coordinates of the center of the line segment from $\cinit$ to $\ce$, and $R$ is the length of the line segment from $\cinit$ to $\ce$. (The diameter of the circular subspace is twice the distance from $\cinit$ to $\ce$.)

We use the OMPL~\citep{sucan_open_2012} implementation of the optimal paths of \cite{reeds_optimal_1990} for the \textsc{Steer} and \textsc{SteerPath} procedures. The distance between the configurations of a path returned by \textsc{SteerPath} is \SI{0.5}{m}.

To find the \textsc{NearestNeighbor} and \textsc{NearNodes}, we use a 3D K-d tree structure.

For collision detection in the \textsc{Collides} and \textsc{RemoveColliding} procedures, we use the GJK algorithm from \emph{cute\_headers} library~\citep{randygaul_cute_headerscute_c2h_2019}.

We terminate the out-of-slot planner at the iteration limit or when the path cost has not improved in five consecutive algorithm resets.

The algorithms are written in C++. We use the \emph{jsoncpp} library to read scenario definitions and write the results.

%% file: eval.tex
\section{Experiments and Evaluation}
\label{s:eval}

In this section we present the results of the algorithm.
We begin with computational experiments.
First, we evaluate the combination of in-slot and out-of-slot planners in simple, randomly generated scenarios in Section~\ref{s:eval:ss}.
Then we continue with real-world scenarios with and without artificial obstacles in Sections~\ref{s:eval:rw} and \ref{s:eval:rwa}.
Finally, we report results of physical experiments with Porsche Cayenne in \cref{sec:real-vehicle-tests}.

The RRT-based out-of-slot planner incorporates the enhancements presented in this paper, i.e., cost heuristic, goal zone, and path optimization, as well as other well-known enhancements, namely, Anytime RRT, K-d tree, and steering toward the goal. To evaluate our contribution, we need to distinguish between these our and well known enhancements. Therefore, we refer to the out-of-slot planner with all the improvements as \emph{OSP-All} and the out-of-slot planner with only the well-known improvements as \emph{OSP-WK}.

For our computational experiments, we use the vehicle dimensions of the Renault Zoe from~\cite{vorobieva_automatic_2015}: $w=\SI{1.625}{m}$, $d_\text{f}=\SI{3.105}{m}$, $d_\text{r}=\SI{0.655}{m}$, $b=\SI{2.450}{m}$, and $\phi_\text{max}=31.4^{\circ}$, which gives the diameter of the turning circle (curb to curb) of the car \SI{10.820}{m}. The maximum number of iterations of the out-of-slot planner is $I_\text{max}=1000$.

We run the computational experiments on a single core of Intel(R) Core(TM) i7-5600U CPU @\SI{2.60}{GHz} with \SI{16}{GB} RAM.

\subsection{Simple Scenarios}
\label{s:eval:ss}

We test our planners in randomly generated scenarios that we call \emph{simple}.
Their parameters are inspired by~\cite{sun_successive_2022} and the algorithm used to generate them is available in our repository\footnote{https://rtime.ciirc.cvut.cz/gitweb/hubacji1/iamcar2.git}.

Each scenario is generated as follows: The initial configuration is $\cinit=(0, 0, 0)$.
The parking slot is placed in such a way that the goal configuration results to $\cinside=(34, 0, 0)$.
Non-overlapping rectangular obstacles are of the same size as the car, i.e., \SI{4.084}{m} long and \SI{1.771}{m} wide.
Cartesian coordinates of obstacle centers and obstacle orientations are generated randomly from the uniform distribution $[10;30]$ and $[-15;15]$ meters and $(-\pi; \pi]$, respectively.
We also add ``wall'' obstacles to limit the state space of the scenario to area of \SI{45}{m}$\times$\SI{30}{m}.

We generate a group of \num{10000} scenarios for each number of obstacles from 1 to 10, giving \num{100000} scenarios in total.
For each scenario, we run our algorithm once and report aggregated results for each group in \cref{t:simple-scenarios}.

\begin{table*}
\centering
\caption{Results of computational experiments with simple scenarios.
Every column corresponds to \num{10000} experiments.}
\begin{tabular}{|r|c|c|c|c|c|}
\hline
Number of obstacles & 1 & 2 & 3 & 4 & 5 \\
\hline
Failure rate [\%] & 0.00 & 0.00 & 0.00 & 0.00 & 0.00 \\
Maximum computation time [ms] & 260 & 263 & 333 & 386 & 365 \\
Average computation time [ms] & 100 & 100 & 102 & 105 & 109 \\
Computation time standard deviation & 20 & 23 & 27 & 31 & 36 \\
Maximum final path cost [m] & 109.83 & 112.32 & 129.59 & 117.83 & 117.73 \\
Average final path cost [m] & 41.01 & 42.33 & 43.90 & 45.54 & 47.55 \\
Final path cost standard deviation & 8.05 & 9.07 & 10.65 & 11.77 & 12.96 \\
\hline
\hline
Number of obstacles & 6 & 7 & 8 & 9 & 10 \\
\hline
Failure rate [\%] & 0.01 & 0.01 & 0.10 & 0.41 & 1.49 \\
Maximum computation time [ms] & 685 & 647 & 783 & 844 & 786 \\
Average computation time [ms] & 114 & 122 & 132 & 145 & 168 \\
Computation time standard deviation & 41 & 51 & 63 & 77 & 101 \\
Maximum final path cost [m] & 125.72 & 123.60 & 143.62 & 148.53 & 147.13 \\
Average final path cost [m] & 49.30 & 51.13 & 53.09 & 55.35 & 56.76 \\
Final path cost standard deviation & 14.06 & 14.62 & 15.90 & 17.08 & 18.68 \\
\hline
\end{tabular}
\label{t:simple-scenarios}
\end{table*}

\begin{figure}[h]
\centering
    \begin{tabular}{cc}
    \includegraphics[width=0.45\linewidth]{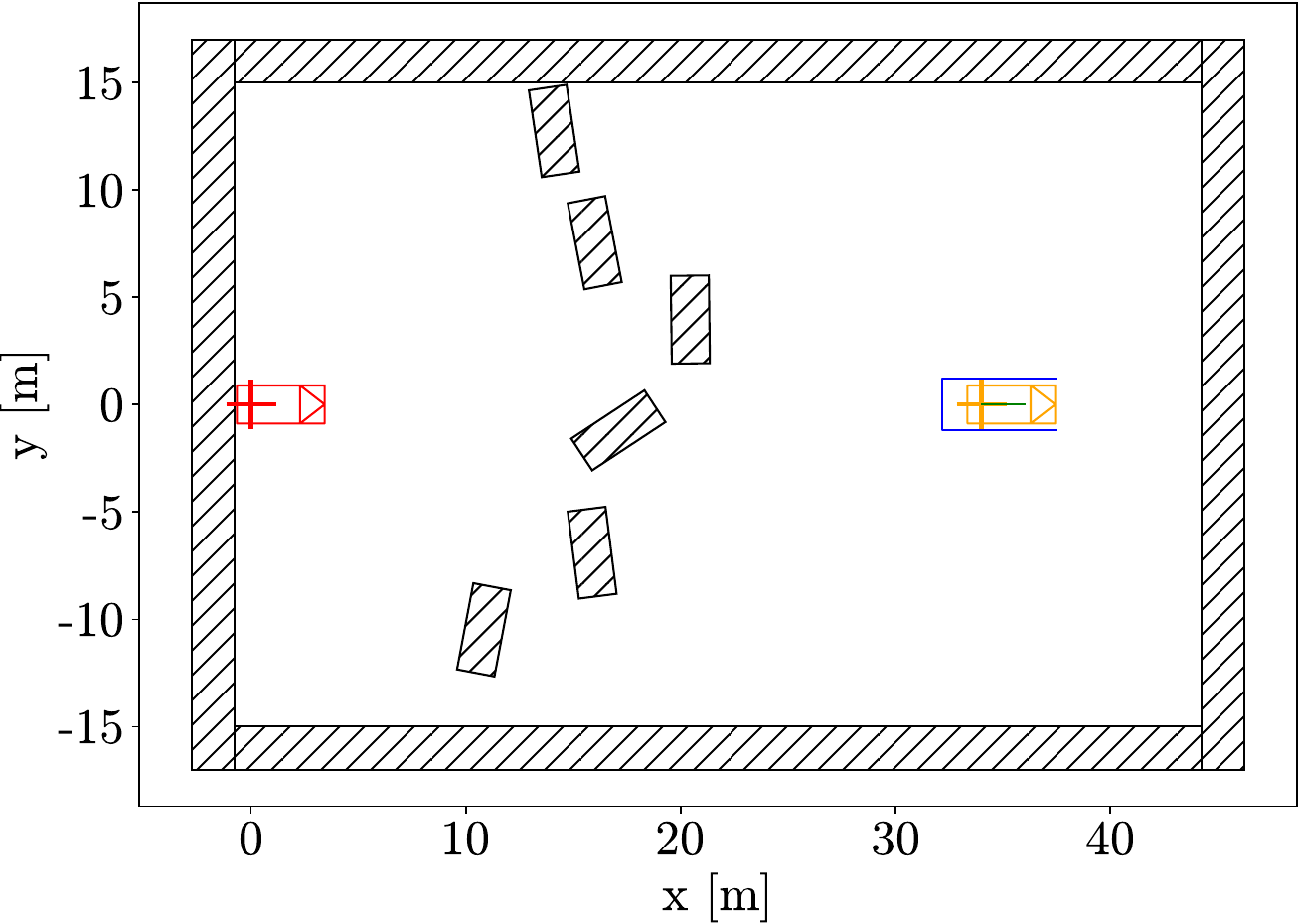} &
    \includegraphics[width=0.45\linewidth]{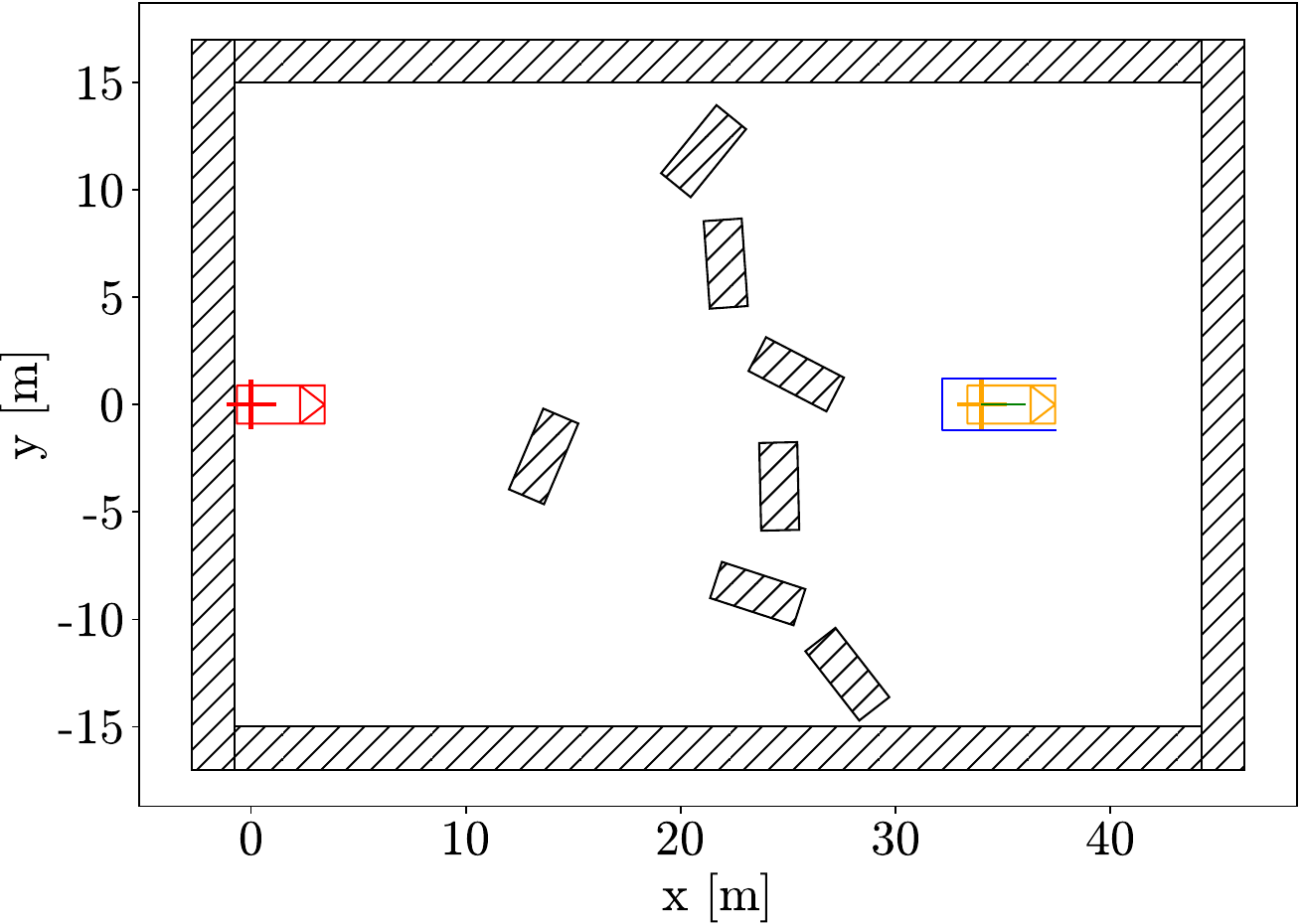} \\
    \end{tabular}
\caption{Selected failing simple parking scenarios for 6 (left) and 7 (right) obstacles. Existence of feasible path is not guaranteed by our scenario generator. The color legend is the same as in \cref{f:intro}.}
\label{f:sps-unsolved}
\end{figure}

We can see that failure rate for scenarios with five and less obstacles is \SI{0}{\%}.
For higher number of obstacles, the rate increases up to the highest value $\approx$\SI{1.5}{\%} for 10~obstacles.
\cref{f:sps-unsolved} shows some failed scenarios with six (left) and seven (right) obstacles, respectively.
Note that our scenario generation algorithm does not guarantee the existence of a feasible path and the shown scenarios either do not have any or it would be difficult to find one even for a human driver.

Furthermore, we see that computation times get longer for more obstacles, but the average computation time for our algorithm is always under \SI{200}{ms}, even for the scenario with 10 obstacles.
Recall that the number of iterations of our algorithm is limited to $I_\text{max}=1000$.
Therefore, the computation time is also limited, although indirectly.

We can also see that the average final path cost and its standard deviation is slowly increasing with the number of obstacles.
This is expected as the layout of obstacles often prevents constructing short paths.

One last detail is that minimum path cost returned by our planners is \SI{36.1}{m} for each scenario.
The reason why the minimum path cost is not \SI{34}{m} as expected follows from two stages of our algorithm.
First, the in-slot planner finds goal and entry position, as can be seen in \cref{f:pe:f}, and path inside the parking slot.
Second, the out-of-slot planner finds a feasible path from initial to entry position with the lowest path cost.
The feasible path from initial position to goal position inside the parking slot is the connection of those two paths.
This behavior makes sense in real-world scenarios, where reaching a parking slot from behind is rare.

\begin{figure}[h]
\centering
    \begin{tabular}{cc}
% 3,56,63,70,80,82,86,99,100,110,126,139,142,147,166,167,195,221,233,278,290,293,402,418,420,431,465,495,501,510,522,544,551,587,591,619,682,711,758,775,783,816,871,877,921,
% 99,147,167,290,418,431,465,501,510,551,591,682,783,816,871,877
% simple: 99,551,591,783,
% more: 510,501,
% obst: 167,
% before GITS review: sps-99, sps-167
    \includegraphics[width=0.46\linewidth]{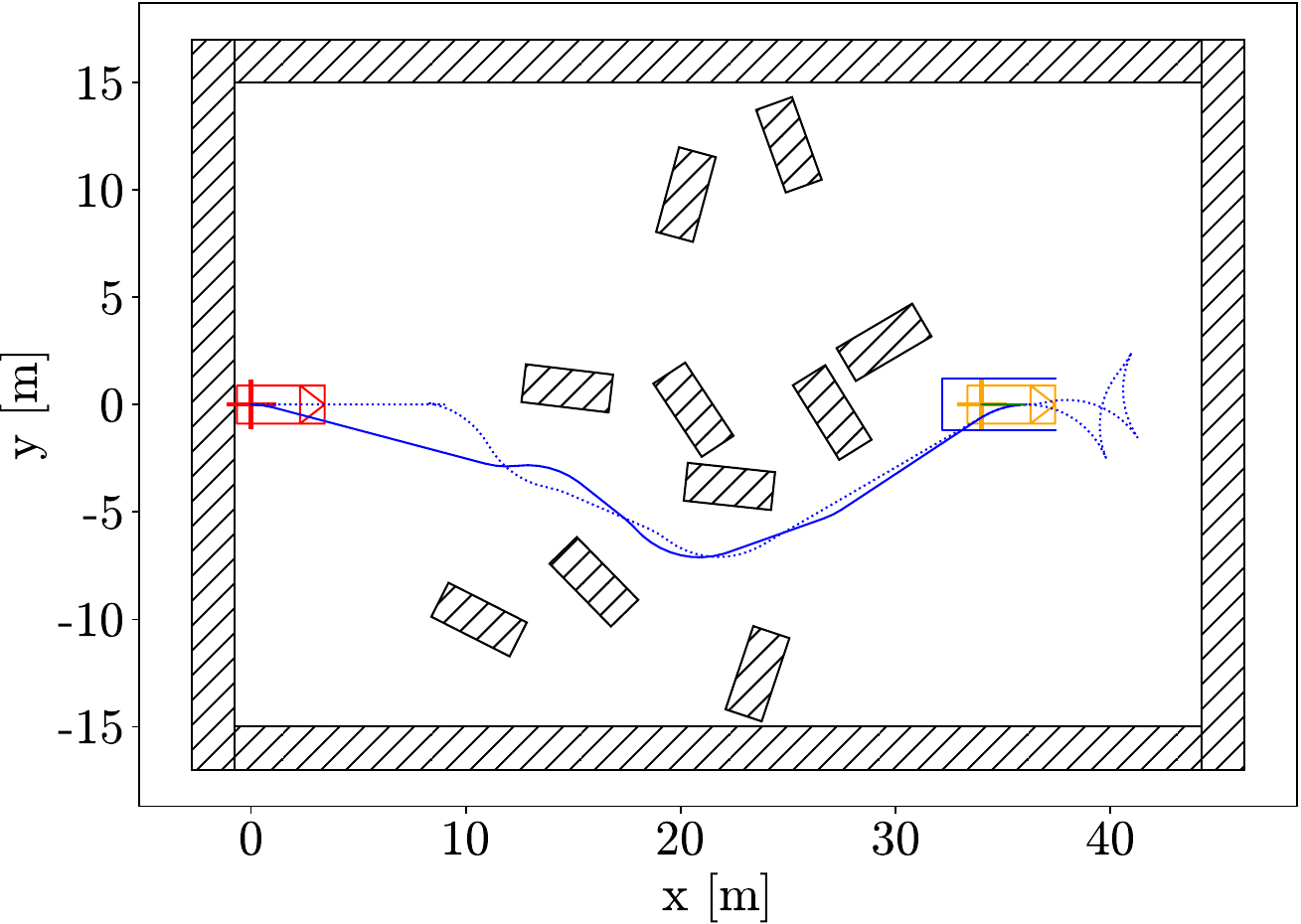} &
    \includegraphics[width=0.46\linewidth]{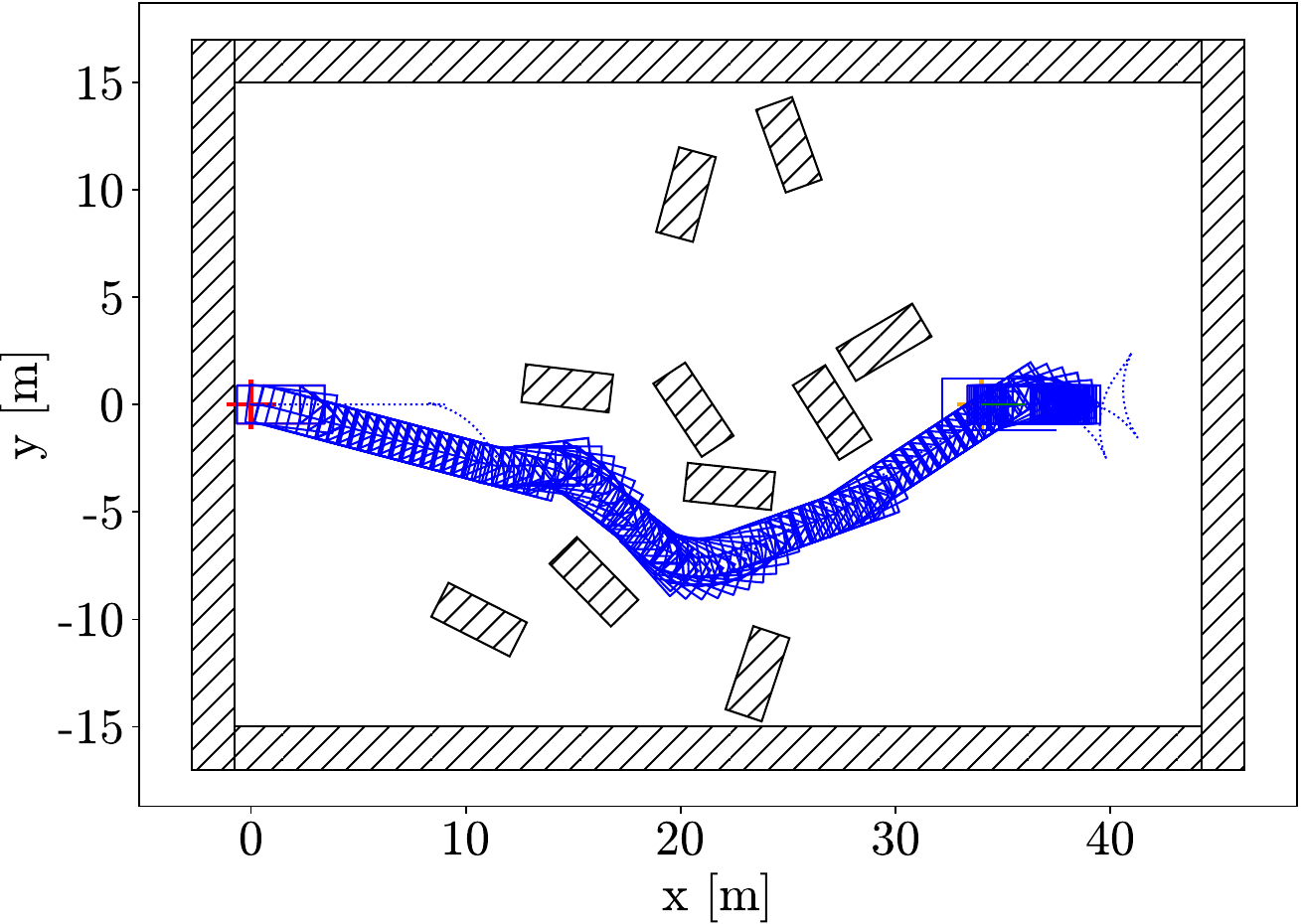} \\
    \end{tabular}
\caption{A simple parking scenario with 10 obstacles. On the left, we can see the idea of the path optimization -- the difference between the solid (optimized) and the dotted (original) path. On the right, there is the car frame visualized along the optimized path. The color legend is the same as in \cref{f:intro}.}
\label{f:sps}
\end{figure}

In \cref{f:sps}, we can see an example of simple parking scenario with 10 obstacles and the solutions of our planners.
Also, we can see the idea of path optimization -- the difference between the solid (optimized) path and the dotted (original) path.
We conclude that the resulting optimized paths are close to paths that would be driven by a human driver.

\subsection{Real-world Scenarios}
\label{s:eval:rw}

\begin{figure}[h]
\centering
    \begin{tabular}{cc}
    \includegraphics[width=0.46\linewidth]{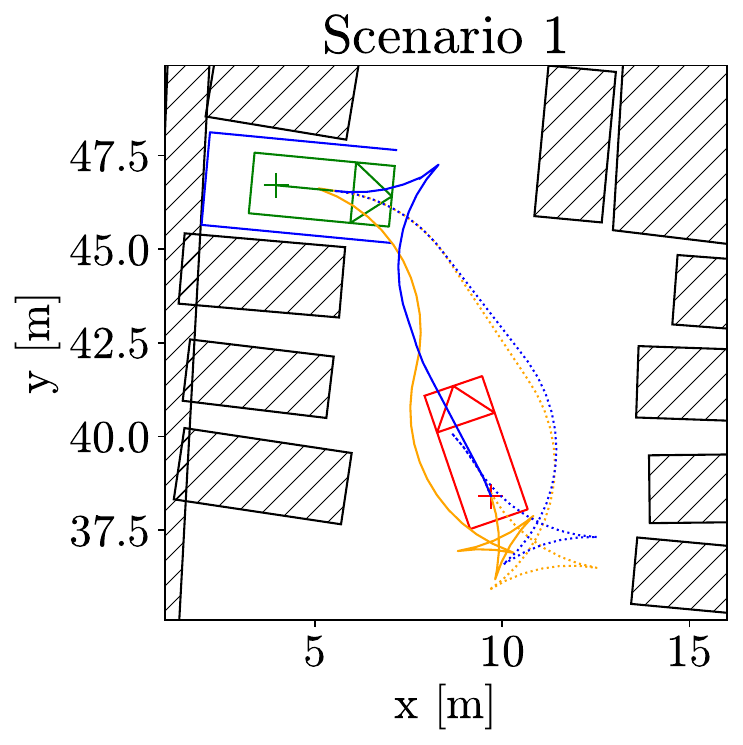} &
    \includegraphics[width=0.46\linewidth]{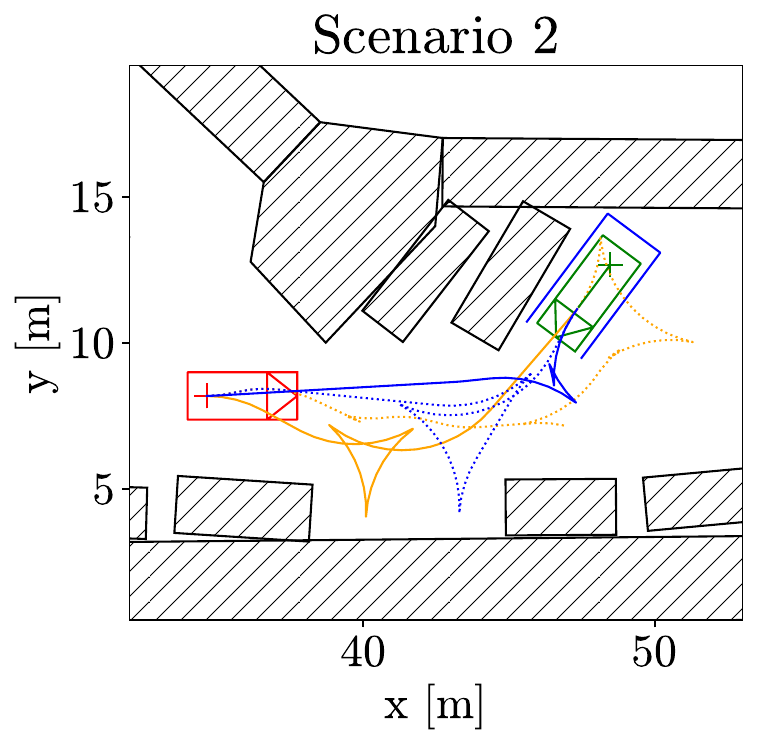} \\
    \includegraphics[width=0.46\linewidth]{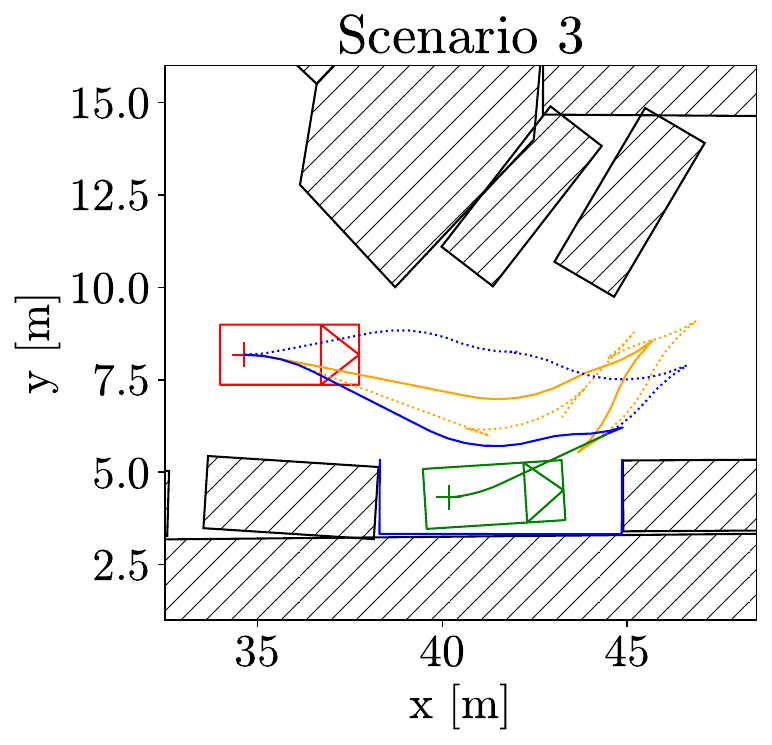} &
    \includegraphics[width=0.46\linewidth]{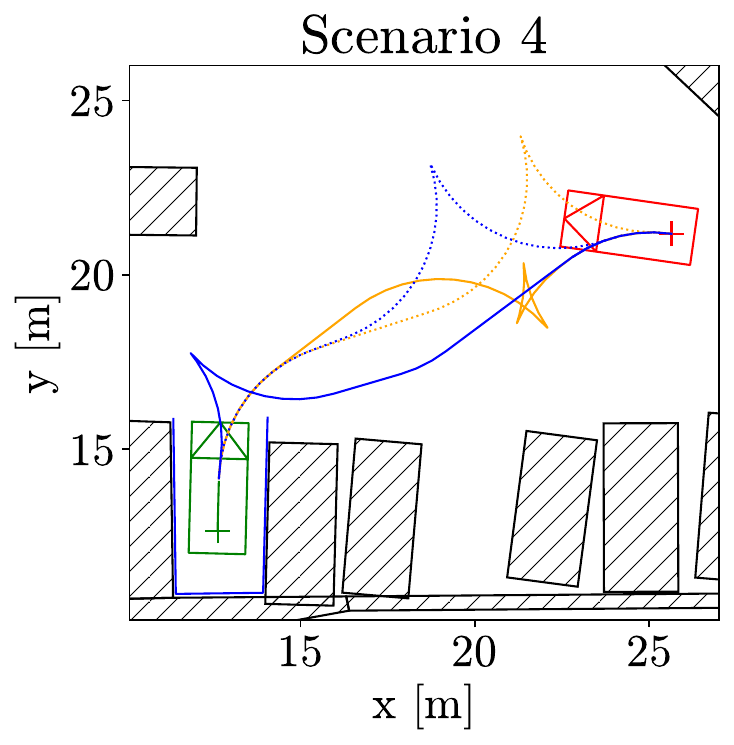}
    \end{tabular}
\caption{Details of the real-world parking scenarios. The color legend is the same as in \cref{f:intro} with the addition of orange lines indicating the worst path found.}
\label{f:rwps}
\end{figure}

We test our planners on four real-world scenarios created from the orthophoto map of Prague streets~\citep{noauthor_ortofoto_nodate}. The detailed layouts of the real-world parking scenarios can be found in \cref{f:rwps}.

We run the out-of-slot planner \SI{10000}{} times for each scenario. The computation times measured on our hardware for the in-slot planner and the out-of-slot planner, as well as the final path costs after optimization, can be found in \cref{t:rwps}.

\begin{table*}[h]
\centering
\caption{Real-world parking scenarios results.}
\begin{tabular}{|r|c|c|c|c|}
\hline
Scenario & 1 & 2 & 3 & 4 \\
\hline
\multicolumn{5}{|c|}{In-slot planner computation time} \\
\hline
Maximum [ms] & $\approx0$ & $\approx0$ & 19 & $\approx0$ \\
\hline
\multicolumn{5}{|c|}{Maximum computation time (plus time of the optimization)} \\
\hline
OSP-All [ms] & 297 + 64 & 920 + 83 & 1 416 + 58 & 452 + 124 \\
OSP-WK [ms] & 3 020 & 8 258 & 22 118 & 1 768 \\
\hline
\multicolumn{5}{|c|}{Average computation time (plus time of the optimization)} \\
\hline
OSP-All [ms] & 131 + 2 & 406 + 8 & 611 + 6 & 209 + 4 \\
OSP-WK [ms] & 685 & 3 310 & 9 762 & 486 \\
Improved by & \textbf{81\%} & \textbf{87\%} & \textbf{94\%} & \textbf{56\%} \\
\hline
\multicolumn{5}{|c|}{Maximum final path cost (after optimization)} \\
\hline
OSP-All [m] & 21.96 & 26.00 & 16.74 & 22.55 \\
OSP-WK [m] & 26.02 & 62.03 & 82.50 & 32.01 \\
\hline
\multicolumn{5}{|c|}{Average final path cost (after optimization)} \\
\hline
OSP-All [m] & 15.93 & 21.88 & 13.78 & 20.12 \\
OSP-WK [m] & 18.91 & 25.89 & 27.09 & 20.99 \\
Improved by & \textbf{16\%} & \textbf{15\%} & \textbf{49\%} & \textbf{4\%} \\
\hline
\multicolumn{5}{|c|}{Minimum number of iterations when a path is found in \SI{100}{\%} of cases} \\
\hline
OSP-All [-] & 59 & 353 & 271 & 43 \\
OSP-WK [-] & 365 & - & - & 120 \\
Improved by & \textbf{84\%} & - & - & \textbf{64\%} \\
\hline
\multicolumn{5}{c}{}
\end{tabular}
\label{t:rwps}
\end{table*}

\begin{figure}[h]
\centering
\includegraphics[width=\linewidth]{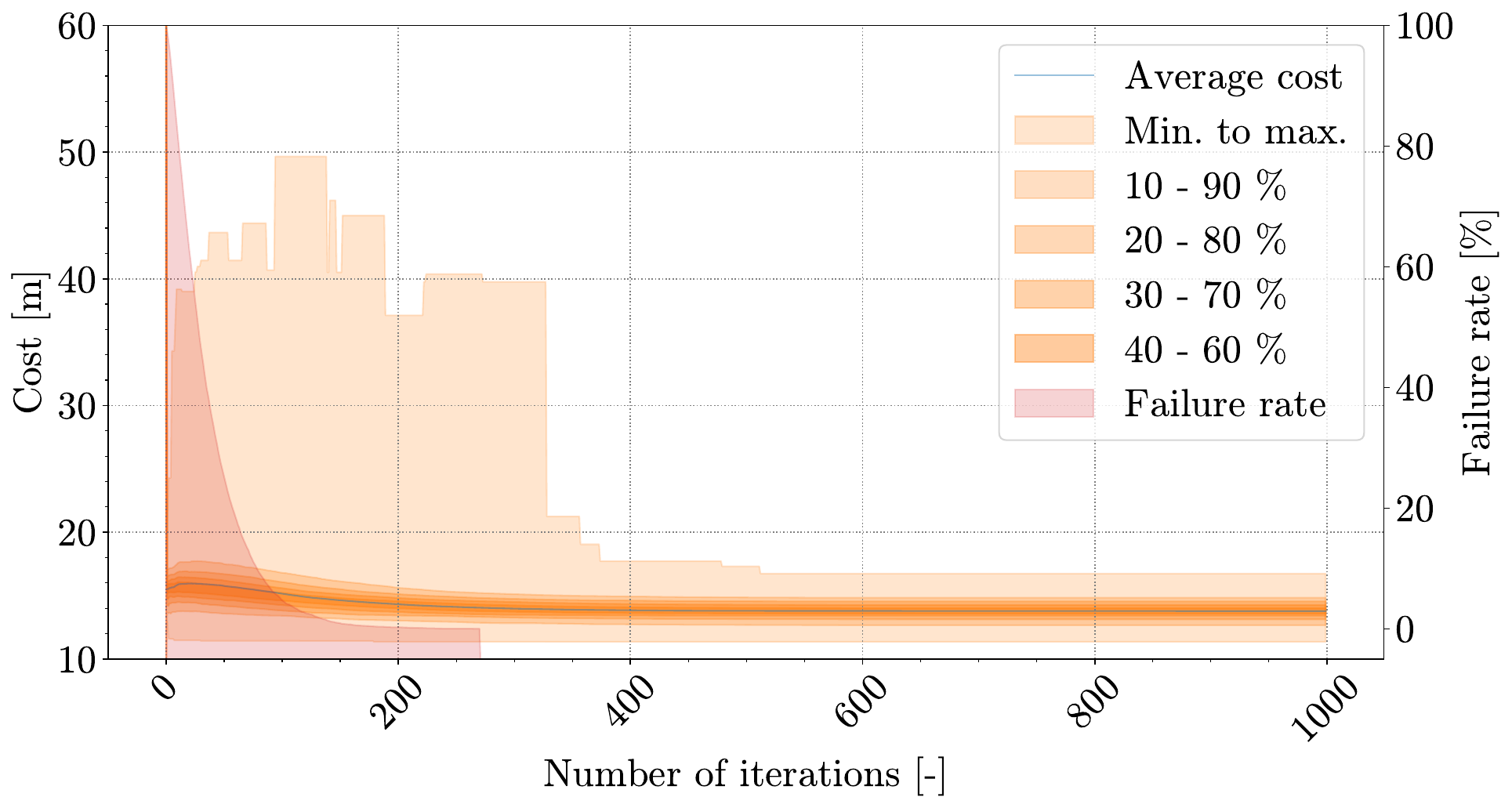}
\caption{Evolution of final path cost after optimization for Scenario~3. The orange area shows the path cost after optimization. The red horizontal line is the average path cost after optimization. The red area shows how many planner runs failed to find the path for the scenario.}
\label{f:rwps-3c}
\end{figure}

\begin{figure}[h]
\centering
\includegraphics[width=\linewidth]{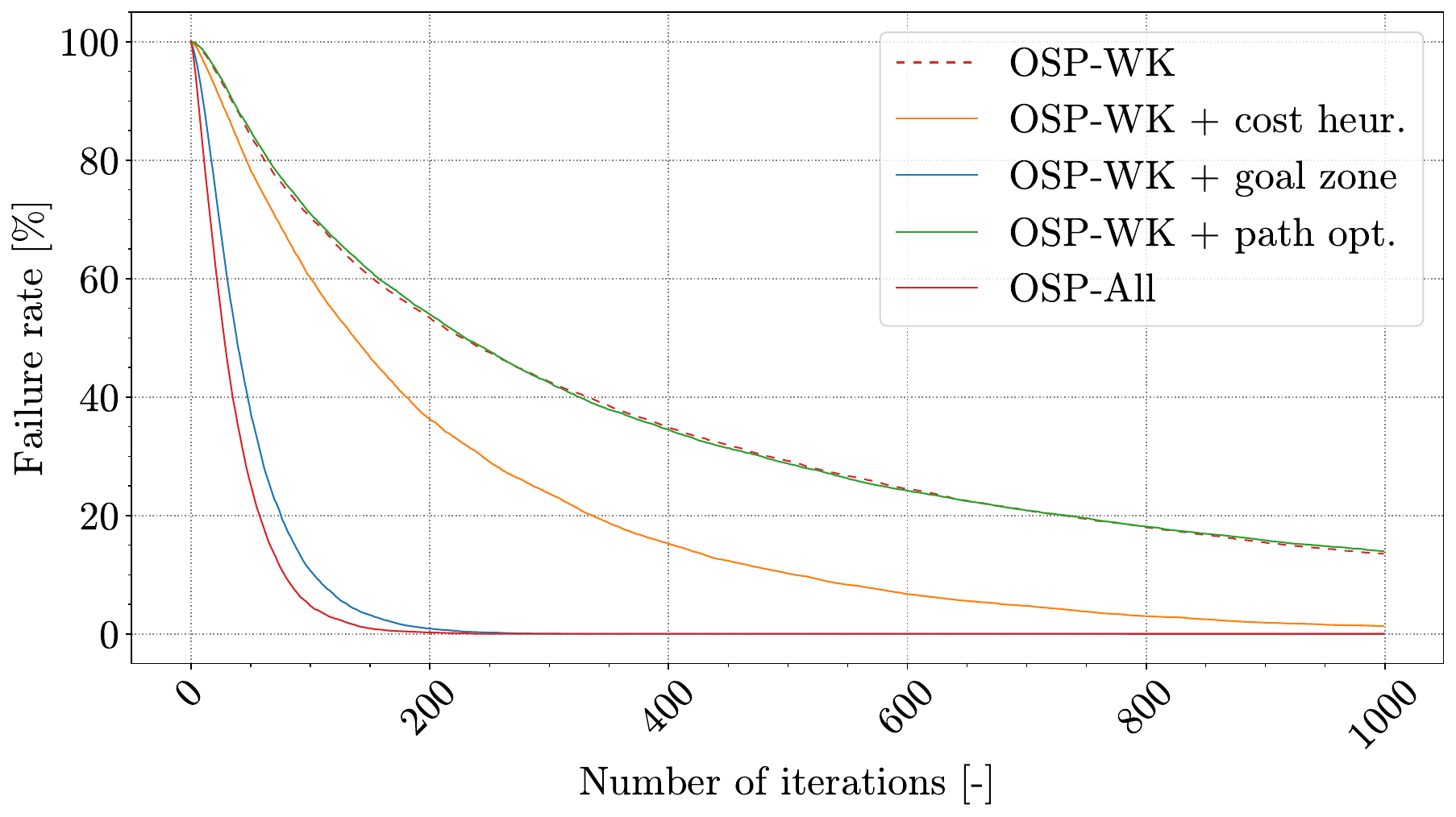}
\caption{Evolution of failure rate for Scenario~3.}
\label{f:rwps-3sr}
\end{figure}

\begin{figure}[h]
\centering
\includegraphics[width=\linewidth]{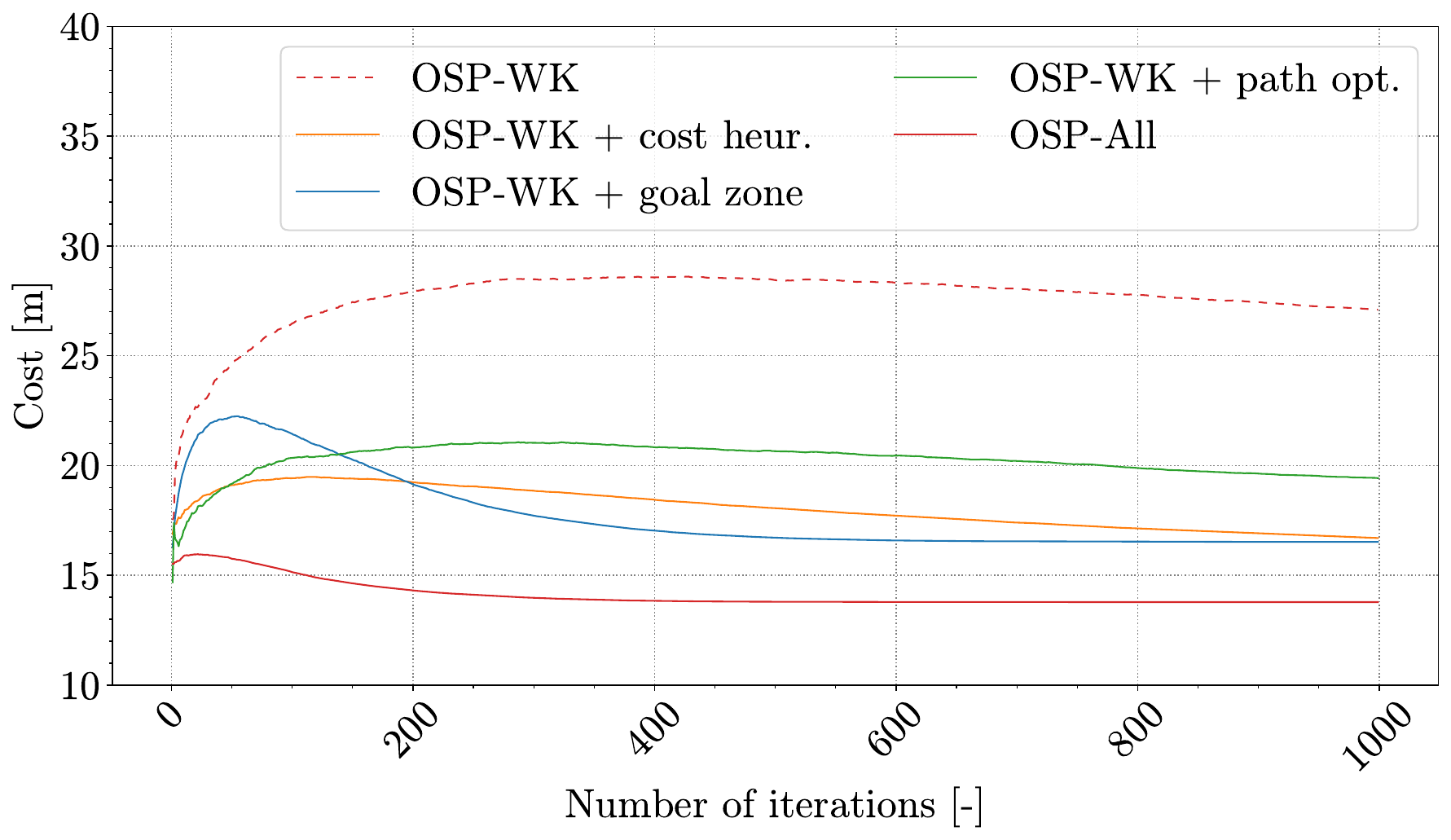}
\caption{Evolution of average final path cost for Scenario~3.}
\label{f:rwps-3ac}
\end{figure}

For more information on how the algorithm works for Scenario~3, the scenario with the longest computation time, see \cref{f:rwps-3c,f:rwps-3sr,f:rwps-3ac}.

\Cref{f:rwps-3c} shows the distribution of the final path cost (including optimization) after each iteration of the algorithm. We see that the cost does not change in the last third of iterations. From the drop of failure rate to zero, we can also see that the path is always found at least after 271 iterations.

In \cref{f:rwps-3sr,f:rwps-3ac}, we show evolution of failure rate and the average final path cost and compare them  for different settings of our out-of-slot planner. There, OSP-WK denotes the results for the out-of-slot planner with only the well-known improvements, i.e., \cref{a:osp} without cost heuristic, goal zone, and path optimization. We also show the results of the planner using only one of our enhancements in isolation, that is: the cost heuristic, goal zone, and path optimization. Finally, we show the results for the out-of-slot planner with all of our enhancements combined, denoted as OSP-All.

In \cref{f:rwps-3sr}, we can see that the cost heuristic and the goal zone  concept lower the failure rate. The cost heuristic lowers the failure rate from \SI{14}{\%} to \SI{1.32}{\%} in the \SI{1000}{}th iteration. The goal zone ensures that a path is found \SI{100}{\%} of the time after at least \SI{399}{} iterations. When used together (OSP-All), a path is found \SI{100}{\%} of the time after at least \SI{271}{} iterations for Scenario~3. For comparison, the failure rate for OSP-WK is \SI{45}{\%} in the same iteration. The path optimization has no influence on the failure rate because it is executed only after the path has been found.

In \cref{f:rwps-3ac}, we can see that the cost heuristic and the goal zone improve the final path cost. This is the secondary effect of the improved failure rate -- a path is found in fewer iterations and the algorithm is reset to find another path that can improve the current best path. As we can see at the last iteration, the cost heuristic improves the final path cost by \SI{38}{\%}, the goal zone by \SI{39}{\%}.
Next, path optimization improves the found path by skipping parts of it, which leads to \SI{28}{\%} enhancements at the last iteration. When combining all enhancements, the average final path cost for Scenario~3 improves by \SI{49}{\%}.

\subsection{Real-world Scenarios with Artificial Obstacles}
\label{s:eval:rwa}

\begin{figure}[h]
\centering
    \begin{tabular}{cc}
    \includegraphics[width=0.46\linewidth]{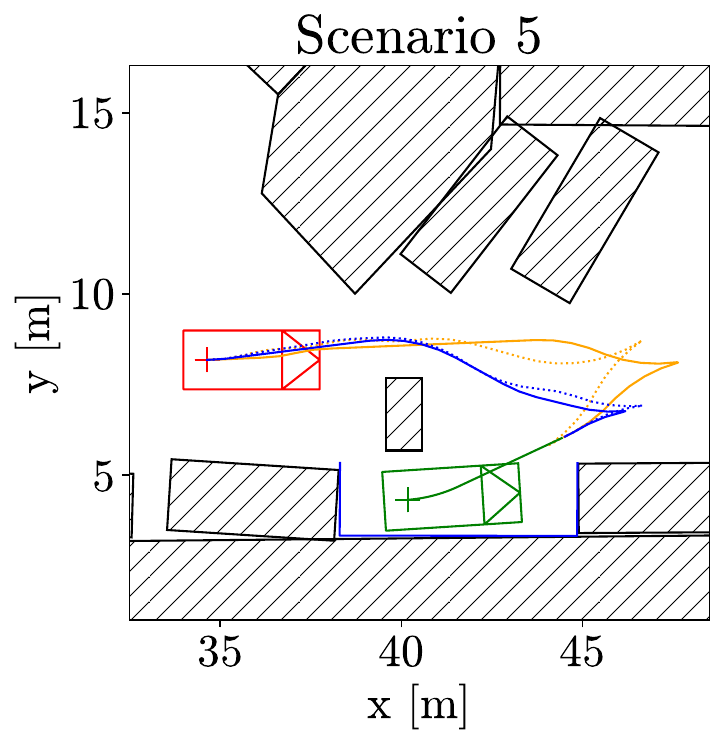} &
    \includegraphics[width=0.46\linewidth]{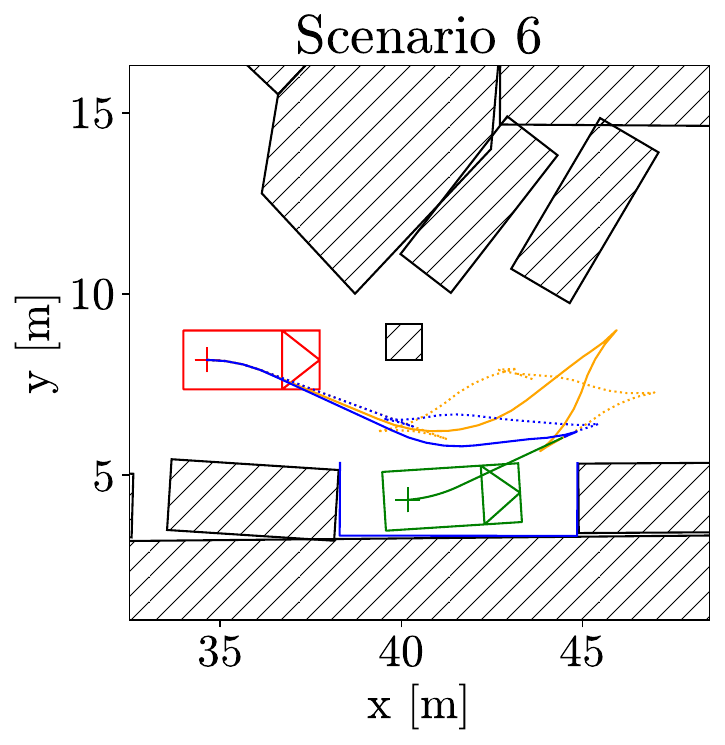} \\
    \includegraphics[width=0.46\linewidth]{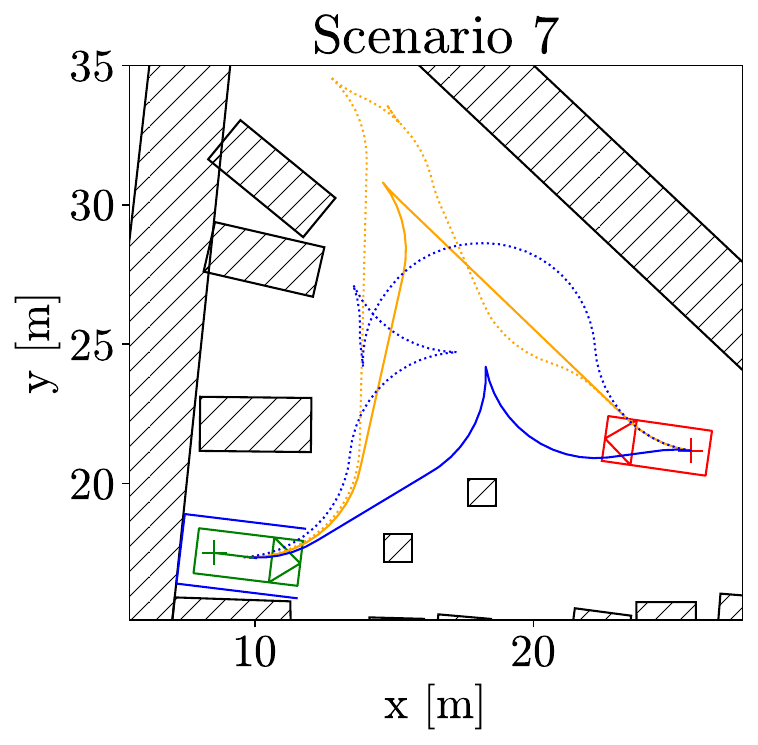} &
    \includegraphics[width=0.46\linewidth]{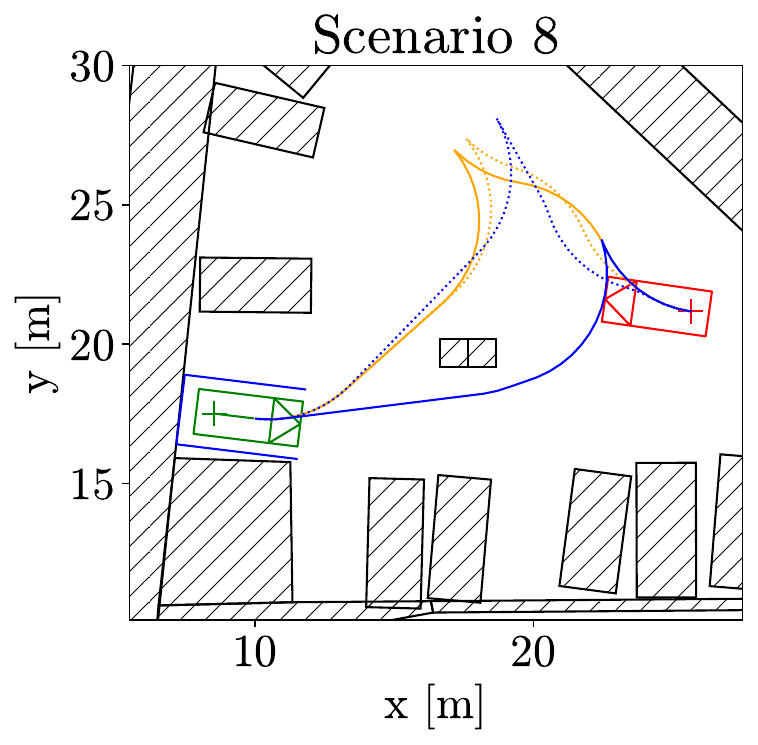}
    \end{tabular}
\caption{Details of the real-world parking scenarios with artificial obstacles. The color legend is the same as in \cref{f:intro}. Additionally, orange indicates the worst path found.}
\label{f:rwps:ao}
\end{figure}

To demonstrate the performance of our planners in constrained environments, we extend the real-world scenarios with artificial obstacles as shown in \cref{f:rwps:ao}.

%To demonstrate the performance of our planners in constrained environments, i.e., the environments, where traditional geometric planners fail, we extend the real-world scenarios with artificial obstacles. \cref{f:rwps:ao} shows the detailed layouts of the extended scenarios and the solutions that are close to the paths that a human driver would choose.

The solutions shown in \cref{f:rwps:ao} are close to the paths that a human driver would choose.
The computation times measured on our hardware for the in-slot planner and the out-of-slot planner, as well as the final path costs after optimization, can be found in \cref{t:rwao}.

\begin{table*}[h]
\centering
\caption{Real-world parking scenarios with artificial obstacles results.}
\begin{tabular}{|r|c|c|c|c|}
\hline
Scenario & 5 & 6 & 7 & 8 \\
\hline
\multicolumn{5}{|c|}{In-slot planner computation time} \\
\hline
Maximum [ms]& 20 & 19 & $\approx0$ & $\approx0$ \\
\hline
\multicolumn{5}{|c|}{Maximum computation time (plus time of the optimization)} \\
\hline
OSP-All [ms]& 885 + 51 & 1 382 + 48 & 2 868 + 712 & 1 244 + 244 \\
OSP-WK [ms]& 8 733 & 11 096 & 10 391 & 4 906 \\
\hline
\multicolumn{5}{|c|}{Average computation time (plus time of the optimization)} \\
\hline
OSP-All [ms]& 421 + 3 & 559 + 5 & 1 372 + 32 & 567 + 32 \\
OSP-WK [ms]& 5 244 & 7 087 & 6 376 & 874 \\
Improved by& \textbf{92\%} & \textbf{92\%} & \textbf{78\%} & \textbf{31\%} \\
\hline
\multicolumn{5}{|c|}{Maximum final path cost (after optimization)} \\
\hline
OSP-All [m] & 17.35 & 17.61 & 31.38 & 25.28 \\
OSP-WK [m] & 75.01 & 72.55 & 76.03 & 30.03 \\
\hline
\multicolumn{5}{|c|}{Average final path cost (after optimization)} \\
\hline
OSP-All [m] & 15.31 & 13.67 & 22.96 & 21.11 \\
OSP-WK [m] & 24.70 & 26.89 & 32.77 & 22.52 \\
Improved by& \textbf{38\%} & \textbf{49\%} & \textbf{30\%} & \textbf{6\%} \\
\hline
\multicolumn{5}{|c|}{Minimum number of iterations when a path is found in \SI{100}{\%} of cases} \\
\hline
OSP-All [-] & 393 & 392 & 464 & 51 \\
OSP-WK [-] & - & - & - & 335 \\
Improved by& - & - & - & \textbf{85\%} \\
\hline
\multicolumn{5}{c}{}
\end{tabular}
\label{t:rwao}
\end{table*}

Scenarios~5 and~6 extend Scenario~3 from \cref{s:eval:rw}. We can see that Scenario 3 requires fewer iterations to find a path with \SI{100}{\%} probability. On the other hand, Scenarios 5 and 6 are solved in less time compared to Scenario 3. This is due to reduction of search space size.

Scenarios~7 and~8 are based on a scenario that, in the absence of artificial obstacles, can be trivially solved using Reeds-Shepp geometric planner. With the addition of the obstacles, the scenarios become more difficult to solve. However, our algorithm finds a good path in an acceptable time of at most \SI{3.5}{s}.

\subsection{Experiments with a Real Vehicle }
\label{sec:real-vehicle-tests}

% textwidth: \printinunitsof{cm}\prntlen{\textwidth}
% linewidth: \printinunitsof{cm}\prntlen{\linewidth}

\begin{figure}[h]
  \centering
  \includegraphics[width=\linewidth]{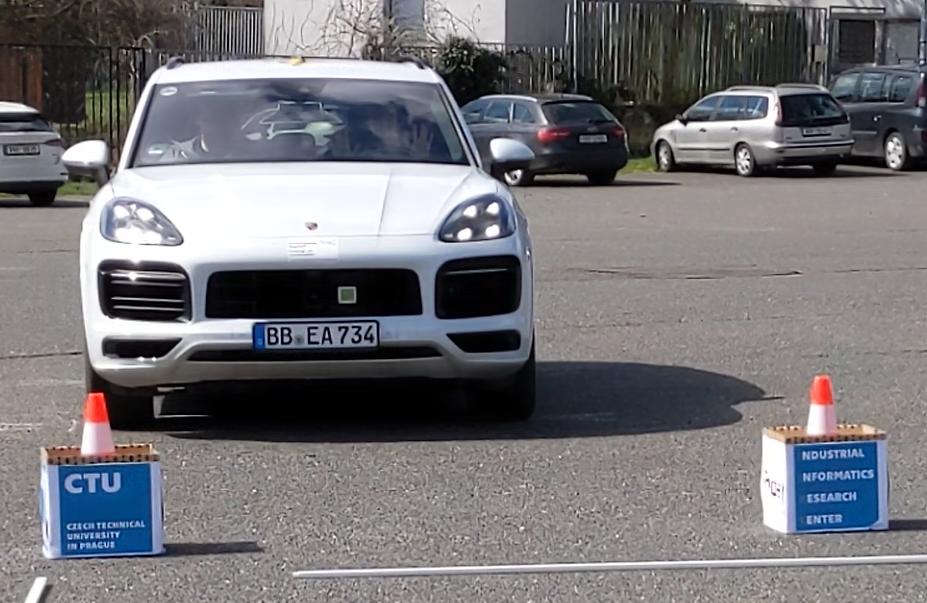}
  \caption{Porsche Cayenne testing the paths produced by our planning algorithm.}
  \label{fig:jupiter}
\end{figure}

To validate our algorithm in practice, we conducted parking experiments with a Porsche Cayenne GTS V8 vehicle (\cref{fig:jupiter}) controlled via FlexRay from our external computer without any obstacle detection system.
We prepared two scenarios with a parking slot and artificially placed obstacles. For each scenario, we manually created a description of the scene that served as input to our planning algorithm and let the algorithm compute the parking path. Then we followed the path in three runs by car and evaluated the difference between the planned and the actual goal configuration. A video of the experiments can be found at \url{https://youtu.be/u_Vqfd5Cn8Q}.

The car was controlled by the software based on the Robot Operating System (ROS) version 2~\citep{ROS}. The path-following algorithm used only signals available from the onboard FlexRay busses. The vehicle was also equipped with an external differential GPS that receives NTRIP corrections and was used only to record the vehicle trajectory for evaluation. The accuracy of the GPS was reported to be less than 1\,cm for position and 1° for heading.

To follow the path of our planning algorithm, we divide it into one or more segments; at the end of each segment, the car stops. The segments end at cusp points where the direction of the path changes, and at inflection points where the sign of the path curvature changes, if both surrounding arcs are longer than 1\,m. Each segment is then followed with trapezoidal speed profile. A path segment is represented by function $\text{seg}:l\rightarrow(v, \phi, \theta)$, which returns the velocity $v$, the steering angle $\phi$, and the expected heading $\theta$ based on the length $l$ from the beginning of the path segment. In the following, the velocity component of this function is abbreviated as $v(l)$ and the other components are abbreviated analogously.

We control the car with two independent controllers running at 50\,Hz, one for longitudinal and one for lateral movements. Both receive the data from the odometry of the car. Specifically, the distance traveled, $l$, is calculated as the cumulative sum of the average speed $v_r$ of the rear wheels, which is reported by the ESP ECU. The heading of the vehicle $\theta$ is made available a signal generated by the EML ECU. Since this signal does not drift in time, we assume that the vehicle computes it by fusing data from odometry and the internal GPS. The longitudinal P controller computes the acceleration demand $a_{\text{dem}} = v(l)' + P_v(v_r-v(l))$, where $v(l)'$ is the time derivative of the reference velocity signal and $P_v$ is the P constant of the controller. The lateral P controller calculates the steering angle demand $\phi_{\text{dem}} = \text{rlim}\left(\phi(l) + P_\theta(\theta - \theta(l))\right)$, where $P_\theta$ is the P constant of the controller and $\text{rlim}$ is a rate limiting function that prevents sending big steps in the demand to the vehicle steering controller.

\begin{figure}[h]
  \centering
  \includegraphics[width=\linewidth]{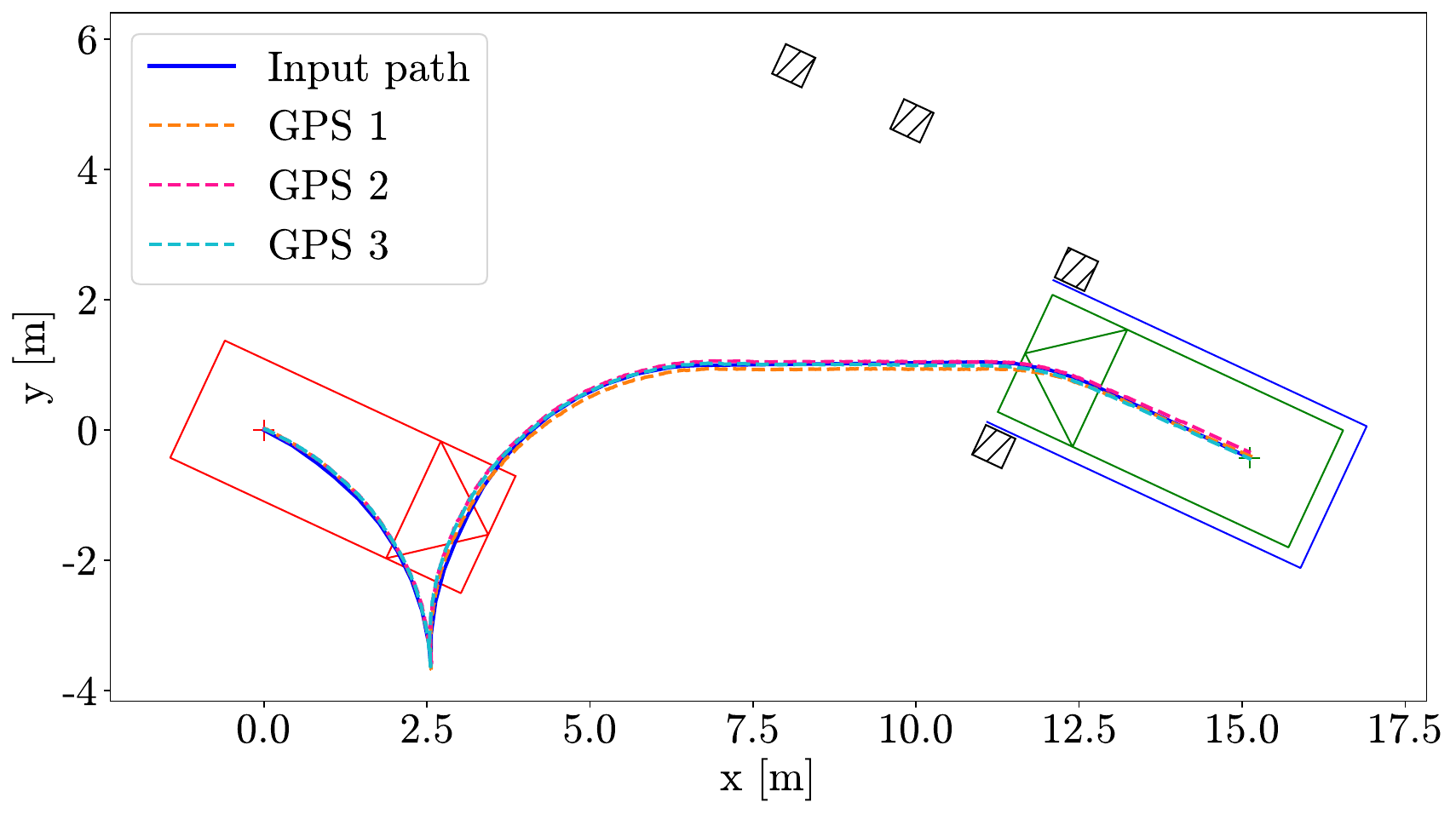}
  \includegraphics[width=\linewidth]{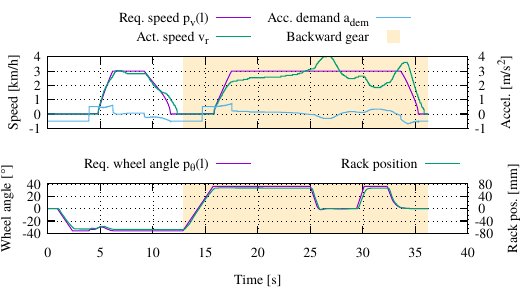}
  \caption{Simple physical experiment --  perpendicular parking with one direction change.}
  \label{fig:drive-pe06}
\end{figure}

\begin{figure}[h]
  \centering
  \includegraphics[width=\linewidth]{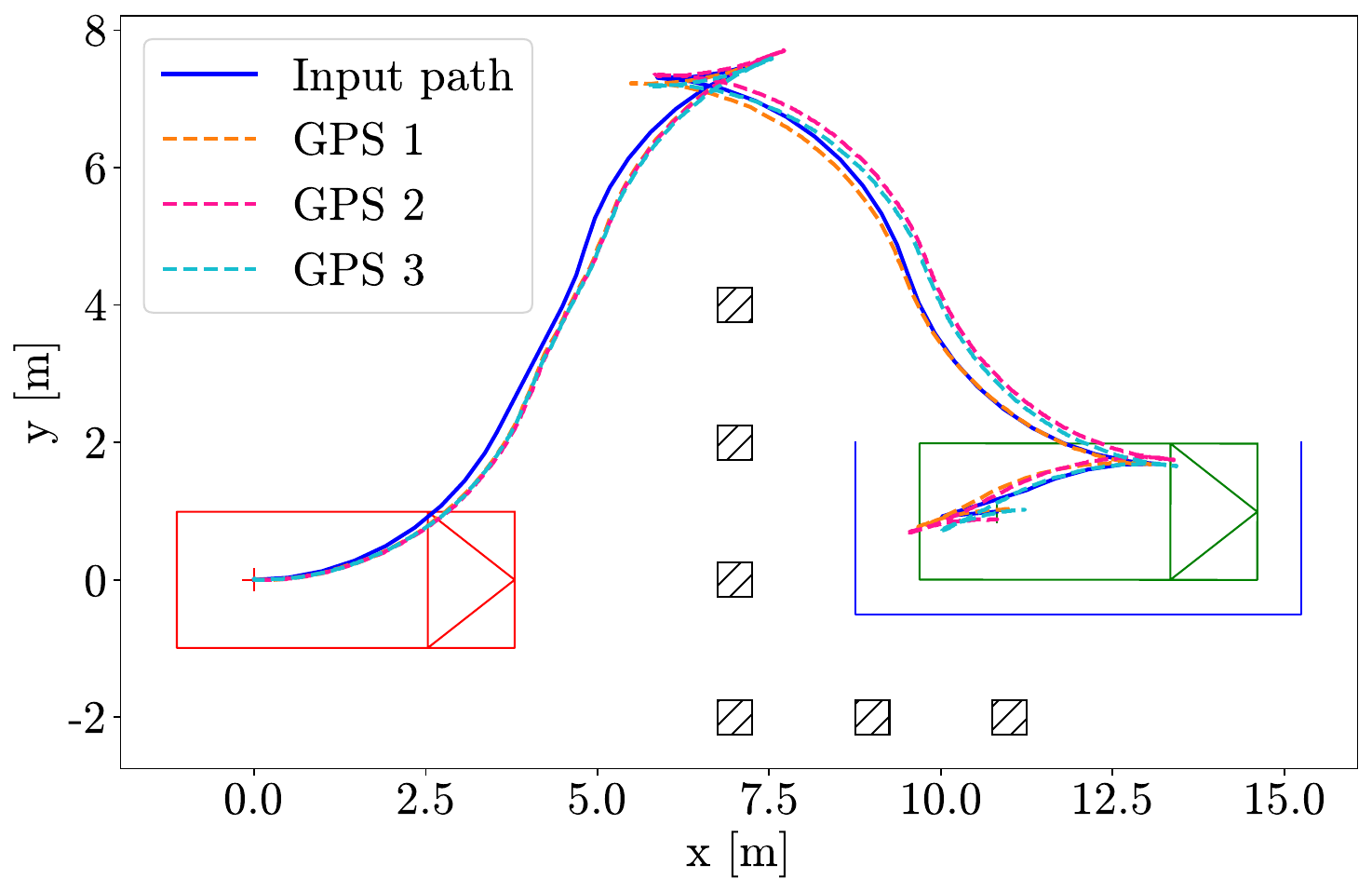}
  \includegraphics[width=\linewidth]{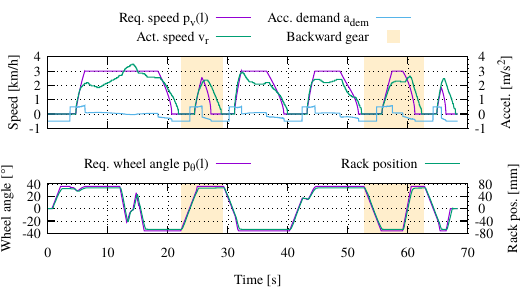}
  \caption{Complex physical experiment -- parallel parking with four direction changes.}
  \label{fig:drive-pa05}
\end{figure}

The data obtained from two scenarios of the physical experiments are shown in \cref{fig:drive-pe06,fig:drive-pa05} with the actual position measured by external GPS in three runs for each scenario.  In the simpler experiment with only one direction change (\cref{fig:drive-pe06}), the vehicle followed the path very closely and the final vehicle positions were always within 10×10\,cm of the planned goal position. In the more complex scenario with four direction changes (\cref{fig:drive-pa05}), the final positions were within 15×40\,cm of the planned goal position. The higher inaccuracy occurs in the longitudinal direction. This is because the car responds to the $a_\text{dem}$ signal with a delay, as can be seen from the velocity plot. We believe that the delay would be shorter and the inaccuracy would be lower if the controller was running in the parking ECU and not in our external computer. Nevertheless, even in this complex scenario, the car was always able to park and avoid obstacles.

